\documentclass[journal]{IEEEtran}

\usepackage[inline]{trackchanges}

\usepackage[hyphens]{url}  
\usepackage{graphicx} 
\usepackage[square,numbers]{natbib}  
\usepackage{caption} 
\usepackage[dvipsnames]{xcolor}
\usepackage{makecell}
\usepackage{amsmath}
\usepackage{amsthm}
\usepackage{algorithm}
\usepackage{algorithmic}
\usepackage{authblk}
\usepackage{lipsum}
\usepackage{amsfonts}

\usepackage{threeparttable}
\usepackage{tabularx}
\usepackage{multirow}
\usepackage{booktabs}
\usepackage{enumerate}
\usepackage{colortbl}
\usepackage{hyperref}
\usepackage{xcolor}
\hypersetup{
	colorlinks,
	linkcolor={red!50!black},
	citecolor={blue!50!black},
	urlcolor={blue!80!black}
}

\title{Few-Sample Traffic Prediction with Graph Networks using Locale as Relational Inductive Biases}

\author{Mingxi~Li,~\IEEEmembership{}
        Yihong~Tang,
        Wei~Ma,~\IEEEmembership{Member, IEEE}
\thanks{The work described in this paper was supported by the National Natural Science Foundation of China (No. 52102385), a grant from the Research Grants Council of the Hong Kong Special Administrative Region, China (Project No. PolyU/25209221), and a grant from the Research Institute for Sustainable Urban Development (RISUD) at the Hong Kong Polytechnic University (Project No. P0038288). \textit{(Corresponding author: Wei Ma.)}}
\thanks{M. Li is with the Department of Civil and Environmental Engineering, The Hong Kong Polytechnic University, Hong Kong SAR, China (E-mail: mingxi-chloe.li@connect.polyu.hk).}
\thanks{Y. Tang is with the Department of Urban Planning and Design (DUPAD), University of Hong Kong, Hong Kong SAR, China (E-mail: yihongt@connect.hku.hk).}
\thanks{W. Ma is with the Department of Civil and Environmental Engineering, The Hong Kong Polytechnic University, Hong Kong SAR, China; The Hong Kong Polytechnic University Shenzhen Research Institute, Shenzhen, Guangdong, China; and Research Institute for Sustainable Urban Development, The Hong Kong Polytechnic University, Hong Kong SAR, China (E-mail: wei.w.ma@polyu.edu.hk).}
}

\begin{document}

\markboth{Manuscript Submitted to IEEE Transactions on Intelligent Transportation Systems, 2022}%
{}

\maketitle
\begin{abstract}

Accurate short-term traffic prediction plays a pivotal role in various smart mobility operation and management systems. Currently, most of the state-of-the-art prediction models are based on graph neural networks (\textsc{Gnn}s), and the required training samples are proportional to the size of the traffic network. In many cities, the available amount of traffic data is substantially below the minimum requirement due to the data collection expense. It is still an open question to develop traffic prediction models with a small size of training data on large-scale networks. We notice that the traffic states of a node for the near future only depend on the traffic states of its localized neighborhoods, which can be represented using the graph relational inductive biases. In view of this, this paper develops a graph network (\textsc{Gn})-based deep learning model \textsc{LocaleGn} that depicts the traffic dynamics using localized data aggregating and updating functions, as well as the node-wise recurrent neural networks. \textsc{LocaleGn} is a light-weighted model designed for training on few samples without over-fitting, and hence it can solve the problem of few-sample traffic prediction. The proposed model is examined on predicting both traffic speed and flow with six datasets, and the experimental results demonstrate that \textsc{LocaleGn} outperforms existing state-of-the-art baseline models. 
It is also demonstrated that the learned knowledge from \textsc{LocaleGn} can be transferred across cities.
The research outcomes can help to develop light-weighted traffic prediction systems, especially for cities lacking historically archived traffic data.  




\end{abstract}


\IEEEpeerreviewmaketitle

\section{Introduction}

\IEEEPARstart{S}{mart} traffic operation and management systems rely on accurate and real-time network-wide traffic prediction \citep{li2021dynamic, meena2020traffic}, and it is the essential input for various smart mobility applications such as personal map services \citep{kaffash2021big}, connected and autonomous vehicles \citep{ma2019trafficpredict}, traffic signal control \citep{zang2020metalight}, and advanced traveler information system/advanced traffic management system (ATIS/ATMS). In many megacities like Los Angeles and New York, massive traffic data have been collected and archived, which include vehicle speeds, traffic volumes, origin-destination (OD) matrices, etc., and these data are widely used to generate traffic predictions. Among different traffic prediction models, deep neural networks, especially the graph-based neural networks, such as graph convolutional networks (\textsc{Gcn}) \citep{zhou2020graph, jiang2021graph}, achieve state-of-the-art accuracy and have been widely deployed in various industry-level smart mobility applications. For example, Uber has been using deep learning for travel time prediction \citep{bell_smyl_2018}.

\begin{figure}[h]
  \center{\includegraphics[width=\linewidth]{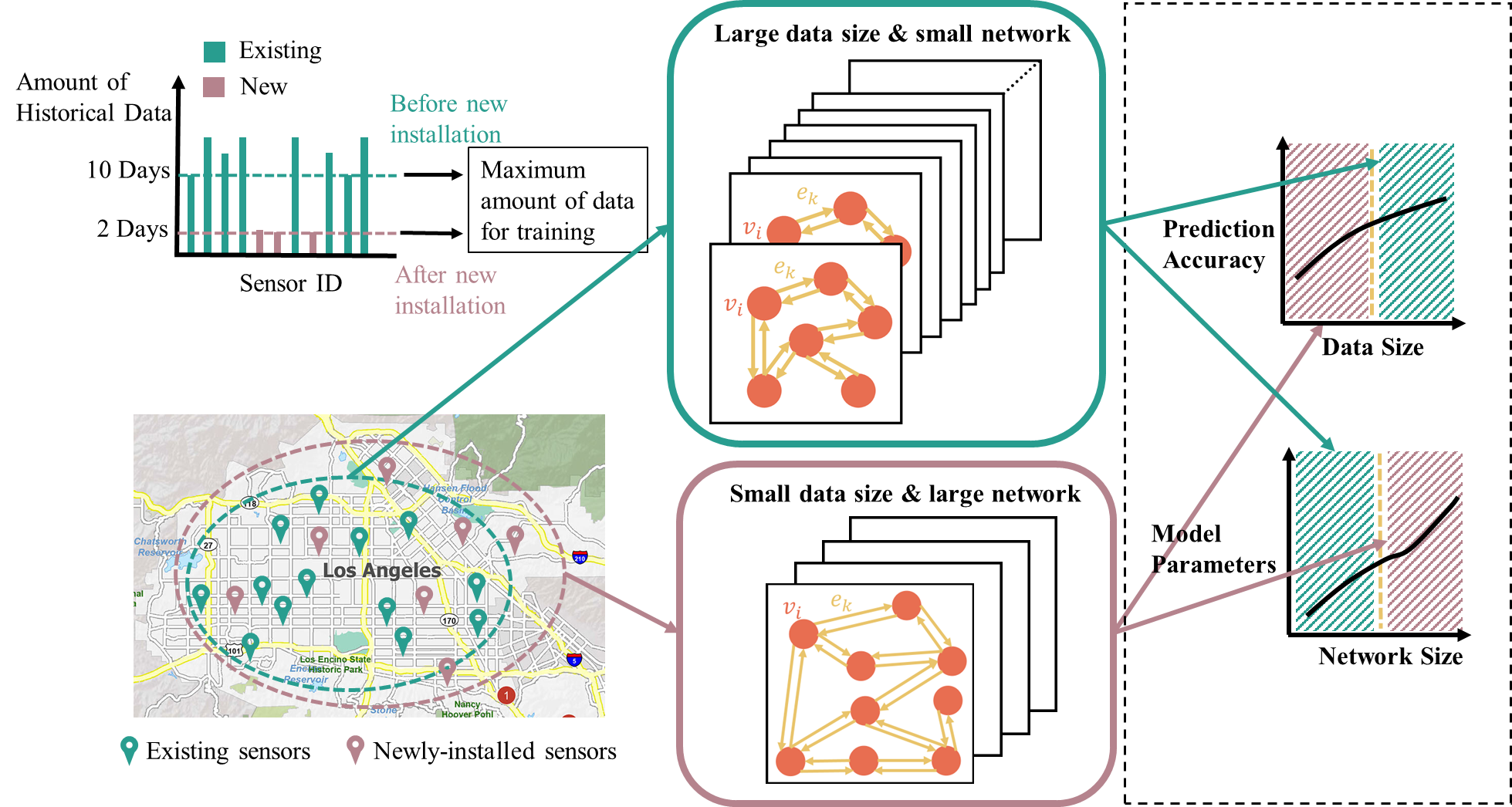}}
  \caption{Relationship of network and data size for current traffic prediction models.}
  \label{fig:relation}
\end{figure}

To further improve the prediction accuracy on large-scale transportation networks, the model complexity and number of trainable parameters of the newly developed traffic prediction models have increased drastically in recent years \citep{yin2021deep, denil2013predicting}. Most of the existing models require the historically archived traffic data for a long time period, and we define the length of the time period as the data size \citep{emmert2020introductory}. In contrast, the size of the transportation network is referred to as the network size. In Figure~\mbox{\ref{fig:relation}}, many cities are gradually installing traffic sensors during the development of the Intelligent Transportation System (ITS). After the new installation, the maximum number of available training data depends on the data amount of the newest sensor. Therefore, the size of the training data for conventional traffic prediction models will be reduced dramatically, when the number of sensors keeps increasing. Every time a new installation of traffic sensors, we need to re-train the traffic prediction model with a smaller dataset and larger graph. For the majority of deep learning models, the number of model parameters is proportional to the road network size. Thus, existing traffic prediction models could overfit on the few training data and the prediction accuracy could drop significantly.


In general, traffic prediction accuracy increases with respect to the data size, and the number of trainable parameters number increases with respect to the network size. 
When the network size becomes larger, more data (longer time period of historical data) is required to prevent overfitting due to a large number of trainable parameters \citep{bartlett2020benign}. 
Overall, existing well-performed deep learning models require a large data size for training to ensure good performance on the large-scale networks. However, because of the high expenses in data collection and sensor maintenance \citep{leduc2008road}, it is not practical to expect every city to archive a comprehensive and long-history dataset. As shown in Figure~\ref{fig:relation}, how to develop traffic prediction models with a small data size on the large-scale networks is the key research question addressed in this paper.

We further motivate this study using a practical example. Hong Kong aims to transform itself into a smart city within the next decade, and the Smart City Blueprint for Hong Kong 2.0 was released in December 2020, which outlines the future smart city applications in Hong Kong. 
The blueprint plans to ``complete the installation of about 1,200 traffic detectors along major roads and all strategic roads to provide additional real-time traffic information'' for Hong Kong's smart mobility system. Consequently, Hong Kong's Transport Department is gradually installing traffic sensors and releasing the data starting from the middle of 2021. The number of traffic sensors increases drastically in the recent year. The duration of the historical traffic data from the new sensors can be less than one month, making it impractical to train existing traffic prediction models. Similar situations also happen in many cities like Paris, Shenzhen, and Liverpool, as the concept of smart cities just steps into the deployment phase globally. Therefore, a network-wide traffic prediction model, which achieves state-of-the-art performance with a small size of traffic data, could enable the smooth transition and early deployment of smart mobility applications. Hence this study has practical values for many cities.

\begin{figure}[h]
    \centering
    \includegraphics[width=.485\textwidth]{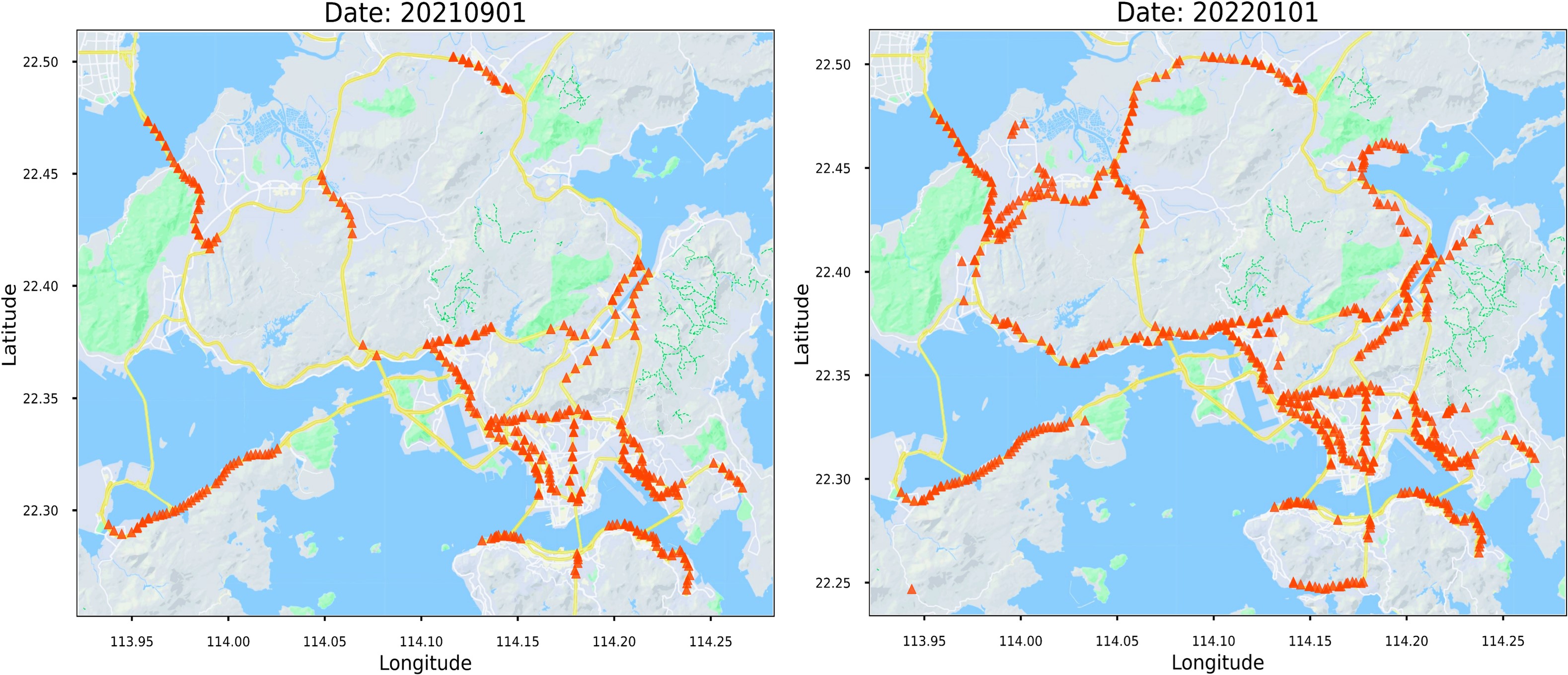}
    \caption{Distribution of available detectors in Hong Kong in September 2021 (left) and January 2022 (right).}
    \label{fig:HKsensor}
\end{figure}

\begin{figure}[h]
  \center{\includegraphics[width=1.0\linewidth]{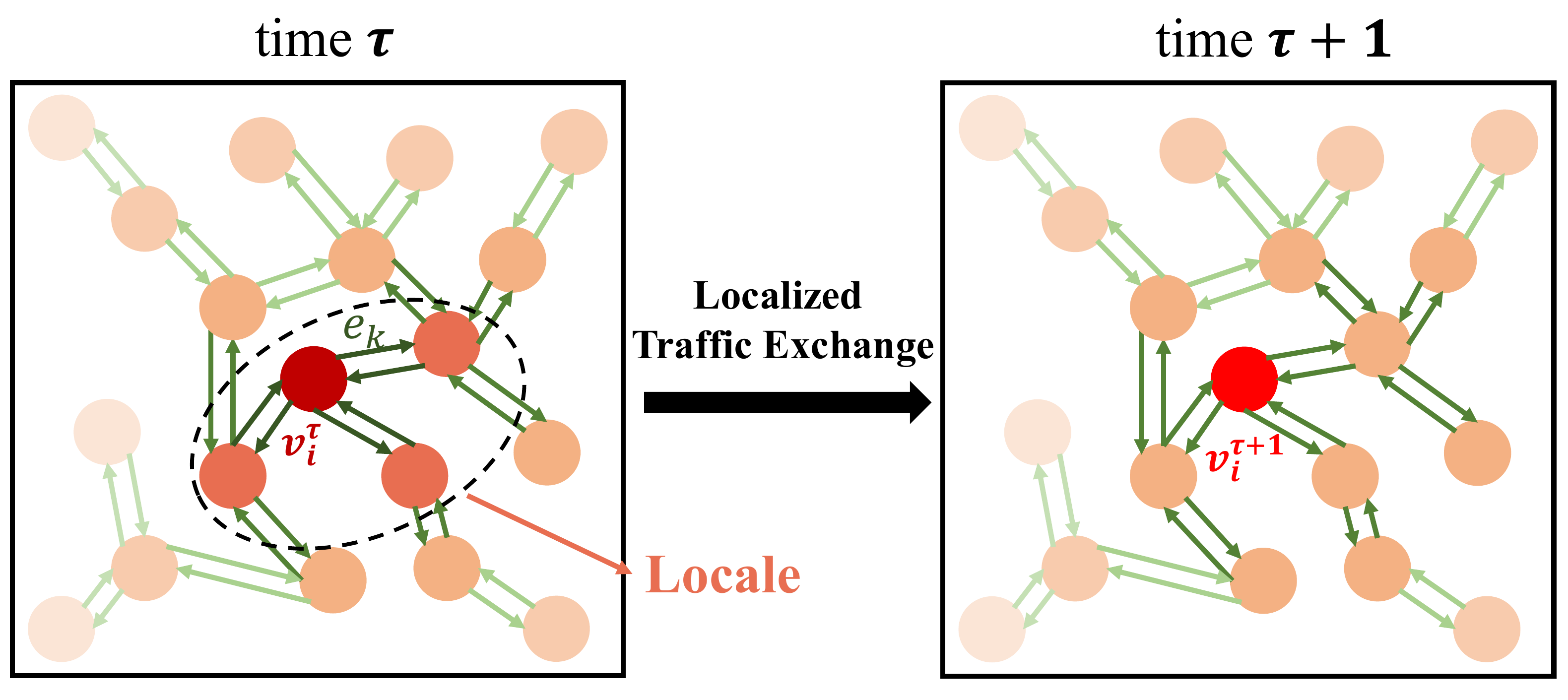}}
  \caption{Illustration of the concept of locale.}
  \label{fig:local}
\end{figure}

To this end, we define the task of few-sample traffic prediction, which aims to train on historical traffic data with a short history, and generate accurate short-term and long-term traffic prediction on the large-scale networks. This task is essentially challenging as the complicated prediction model on large-scale network is prone to overfit with limited data \citep{caruana2001overfitting, ghasemian2019evaluating}. To address this issue, we notice that the short-term traffic state on a certain node (target node) only depends on the traffic states of its localized neighbors. More specifically, as shown in Figure~\ref{fig:local}, the traffic state at time $\tau + 1$ is mainly affected by the traffic states of itself and its neighborhoods at time $\tau$. It is straightforward to observe that the change of traffic state on a node is attributed to the traffic ({\em e.g.,} vehicle, flow) exchanges, then the nodes far away from the target node cannot exchange traffic directly with the target node, and hence the impact of those nodes are indirect and marginal. We define the concept \textbf{locale} of a target node as a collection of information on its neighboring nodes, and the information includes, but is not limited to traffic states ({\em e.g.,} speed, flow, OD), static data ({\em e.g.,} road type, speed limit), and auxiliary information ({\em e.g.}, weather). Finally, it is safe to claim that the traffic state of the target node in the near future mainly depends on its current locale.  



The locale of a node can be viewed as the relational inductive biases for the prediction model, which enforces the entity relations in a deep learning architecture. In our case, the connections between nodes allow the traffic exchanges, while no direct traffic exchanges are allowed if two nodes are not connected. 
Additionally, other static information such as location information and road properties affect the speed and frequency of those traffic exchanges. To make use of the relational inductive biases, we adopt the Graph Network (\textsc{Gn}) to capture the dynamic and localized traffic states of a target node. The \textsc{Gn} demonstrates great potential in modeling the relational inductive biases and has been widely used to depict the localized relationship among different entities \citep{battaglia2018relational}. It is also a generalized form of many existing graph-based network structures \citep{seo2019differentiable,sanchez2018graph}.  In this paper, we extend the \textsc{Gn} model to learn the \textit{locally spatial} and temporal patterns of traffic states for generating predictions. 
Importantly, \textsc{Gn} is applied to each node separately, and it can be applied to different nodes with various topologies. \textsc{Gn} is also light-weighted with a small number of trainable parameters, making it easy to train given the limited training samples.

It is noteworthy that the similar concept of locale is widely adopted for the spatial approaches of the graph neural networks (\textsc{Gnn}) defined in \citep{zhou2020graph}, and one of the representative examples is the message passing neural network (\textsc{Mpnn}) \citep{gilmer2017neural}. However, current traffic prediction models overlook the properties of locale and rely on \textsc{Gnn} to implicitly learn the locally spatial dependency of the traffic states, which also explains why the existing traffic prediction models generally require a large amount of data. To the best of our knowledge, among the commonly used traffic prediction models, \textsc{Dcrnn} and \textsc{Graph WaveNet} depict the diffusion process of traffic states, which share the most similar idea to locale \citep{li2018diffusion, wu2019graph}. However, due to their complex model structures, the performance of both models degrades drastically with a few data samples, which will later be shown in the numerical experiments. This observation also suggests a light-weighted structure for the few-sample traffic prediction task.


To summarize, the small data size and large network size present great challenges to existing traffic prediction models, and it is promising to utilize the localized traffic information and data (locale) to generate traffic predictions without modeling the entire network simultaneously. In view of this, this paper proposes Localized Temporal Graph Network (\textsc{LocaleGn}) to model the \textit{locally spatial} and temporal dependencies on graphs for predicting traffic states in the near future. 
This is the first time that \textsc{Gn} is adopted and extended for traffic prediction tasks, and the choice of \textsc{Gn} is attributed to its previous successful applications in predicting physical systems, such as fluid dynamics and glassy systems \citep{battaglia2018relational}. 
\textsc{LocaleGn} can learn from different nodes on the large network, and hence it can be declared as a light-weighted model to make accurate short-term predictions from few training samples.
The contribution of this paper can be summarized as follows:
\begin{itemize}
\item We propose the few-sample traffic prediction task and highlight the importance of using localized information (locale) of each node to address the issue of lacking data in the proposed task.
\item We develop a {\mbox{\em \textit{locally spatial}}} and temporal graph architecture \textsc{LocaleGn} for the few-sample traffic prediction task. The \textit{locally spatial} pattern is modeled by the graph network with relational inductive biases, and the temporal pattern is depicted with a modified recurrent neural network.  
\item We conduct experiments on three real-world traffic speed datasets and three real-world traffic flow datasets, respectively. The results show that \textsc{LocaleGn} consistently outperforms other state-of-the-art baseline models for the few-sample traffic prediction. It is also demonstrated that the learned knowledge from \textsc{LocaleGn} can be transferred across cities. 
\end{itemize}
Source codes of \textsc{LocaleGn} are publicly available at \url{https://github.com/MingxiLii/LocaleGN}.
The remainder of this paper is organized as follows. Section \ref{sec:literature} reviews the related studies on traffic prediction, few-sample learning, and graph neural networks.  
Section~\ref{sec:model} formulates the task of few-sample traffic prediction and presents details of \textsc{LocaleGn}. In Sections~\ref{sec:exp}, numerical experiments are conducted to demosntrate the advantages of \textsc{LocaleGn}. Lastly, conclusions and future research are summarized in Section \ref{sec:con}.

\section{Related Work}
\label{sec:literature}
In this section, we first summarize the existing traffic prediction models, then applications of few-sample learning in transportation are discussed. Lastly, we review the emergence of \textsc{Gn} and justify the choice of \textsc{Gn} for the few-sample traffic prediction.
\subsubsection{Traffic Prediction Models}
Traffic data as a spatio-temporal information on road networks (graphs) is a typical non-Euclidean data. There are two types of models to deal with graph data: spatial graph-based models and spectral graph-based models \citep{zhou2020graph}. In smart mobility applications, \textsc{Gcn} is a widely used spectral graph-based model to learn the complex structure of spatio-temporal data, and the representative models include \textsc{T-Gcn} \citep{8809901}, \textsc{St-Gcn} \citep{yu2018spatio}, \textsc{Tgc-Lstm} \citep{8917706}, \mbox{\textsc{stemGNN}} \mbox{\citep{cao2020spectral}} and \textsc{AstGcn} \citep{guo2019attention}. The convolutional operation bases on the whole graph with its Laplacian matrix or other variants, such as dynamic Laplacian matrix \citep{diao2019dynamic}.
In \mbox{\citep{peng2020spatial}}, it is a delicate prediction model for passenger inflow and outflow at traffic stations, such as subway or bus station. In other words, the graph consists of nodes (traffic stations) with two attributes (inflow and outflow). The graph information is further processed to obtain the subgraph for each node. This is an enlightening setting, in which spatial-temporal patterns can be learned by subgraph cantered with one node. 
However, the spectral graph-based models do not have parameter-sharing characteristics, thus these models are tied to specific graph topologies. For example, \textsc{T-Gcn} and \textsc{St-Gcn} are inflexible and impose restrictions on transfer learning tasks across different graphs \citep{scarselli2008graph}.  

For spatial graph-based models, local aggregators are explored in order to deal with the direct convolution operation with the different number of neighbors \citep{zhou2020graph}. In \textsc{Dcrnn} \citep{li2018diffusion}, the bidirectional random walk is introduced to model the spatial correlation of traffic flow on graphs. In \textsc{Graph WaveNet}, an adaptive dependency matrix is adopted through the node embedding process, and the 1D convolution component is used to make the relation between entities trainable \citep{wu2019graph}. Moreover, attention-based spatial methods are developed to operate the attention mechanism on graphs. In \textsc{Gatcn}, the graph attention network is applied to extract the spatial features of the road network \citep{guo2020short}. We note that existing spatial graph-based models have flexible graph representations but require large historical data to achieve competitive accuracy. For example, in \textsc{Dcrnn}, the pair-wise spatial correlation makes graph structures trainable, but its complex encoder and decoder structures also make the training data-intensive. 

Methods mentioned above in the spectral and spatial domain mainly focus on modeling the spatial dependency of the traffic data. In addition to capturing spatial patterns, many other models are combined to learn temporal patterns. For example, Recurrent Neural Network (\textsc{Rnn}) is adopted in time series prediction tasks \citep{lipton2015critical, yu2019review}. Both \textsc{Lstm} and \textsc{Gru} are two typical \textsc{Rnn} models \citep{tian2015predicting}, and attention mechanism can also be used for inferring temporal dependencies ({\em e.g.,} \textsc{Ast-gat} \citep{li2021spatiotemporal}, attention with \textsc{Lstm} \citep{zheng2020hybrid}, and \textsc{Aptn} \citep{shi2020spatial}).

In addition to deep learning, there are also other methods to deal with traffic prediction. For example, in \mbox{\cite{peng2021dynamic}}, reinforcement learning is utilized to address data defect issues in traffic flow prediction. The dynamic graph is generated by the policy gradient algorithm, and the differences of the graph are used as reward signals for reinforcement learning to generate action sequences on the traffic flow transfer graph. The linear models have more compact and simple structures, which are used in many traffic prediction settings. For example, \mbox{\cite{YILDIRIMOGLU201345}} proposes to predict the short-term traffic dynamics based on the shockwave analysis and bottleneck identification. Specifically, the historical dataset is clustered by Gaussian Mixture Model and the clustering result is used for generating the congestion map bounded by shockwave speed. Besides, \mbox{\cite{kwak2020travel}} develops a dynamic linear model with recursive coefficient updates.  On top of \mbox{\cite{kwak2020travel}}, the large-scale dynamic linear model (\textsc{Lsdlm}) further incorporates graph topological information and makes it feasible to make prediction on large scale network \mbox{\cite{kwak2021traffic}}. In general, it is difficult to adapt the above models for the traffic prediction settings with changing the number of sensors.

\subsubsection{Few-shot and Few-sample Learning in Transportation}
The majority of research applying deep learning to small datasets focuses on few-shot prediction which includes pre-training on a large dataset and fine-tuning on a small dataset \mbox{\citep{lin2018transfer, zhang2019deep, 8880220}}. However, works for directly learning from small datasets without external data are scarce \mbox{\citep{barz2020deep}}. Few-sample learning can be viewed as a specific example of few-shot learning without pre-training. Thus, it is a promising methodology to deal with the data inefficiency issue in the field of transportation. There is no direct work about few-sample learning in transportation, but some few-shot learning approaches are explored. The existing few-shot learning models can be categorized into two types: one is the gradient-based models \mbox{\citep{finn2017model}} and the other is the metric-based models \mbox{\citep{snell2017prototypical}}. These methodologies work well in the case of independent and identically distributed data. However, few-shot learning in graphs is more challenging as the nodes and edges on the graph are connected and correlated with each other. Transportation data is a typical type of graph-based data. In recent years, studies explore few-shot learning with transportation applications such as vehicle detection \mbox{\citep{cao2016towards}} and traffic sign recognition \mbox{\citep{lin2019transfer}}. In traffic prediction tasks, region-based transfer learning across cities is applied by matching similar sub-regions among different cities \mbox{\citep{Wang2019CrossCityTL}}. Overall, 
 although lacking few-sample learning, the explorations of few-shot learning for traffic prediction can provide inspiration for building few-sample models without external data.

\subsubsection{Graph Networks}
Spectral \textsc{Gcn} models and spatial \textsc{Gnn} models can be adopted for traffic prediction tasks \citep{jiang2021graph}. The inflexibility of the former one makes it unfeasible to construct few-sample or transfer prediction models. Therefore, it is worthwhile to look into spatial-based \textsc{Gnn} models. There are many popular spatial-based \textsc{Gnn} models such as diffusion graph convolution (\textsc{Dgc}) \citep{atwood2016diffusion}, \textsc{Mpnn} \citep{gilmer2017neural}, \textsc{GraphSAGE} \citep{hamilton2017inductive}, and graph attention networks (\textsc{Gat}) \citep{velickovic2017graph}. 
Traffic speed or flow prediction models base on \textsc{Dgc}, \textsc{Mpnn}, \textsc{GraphSAGE} and \textsc{Gat} for short-term prediction are developed with sufficient traffic data. However, these models may fail in the cases of few data samples. In view of this, we note that Graph Network (\textsc{Gn}) with relational inductive biases could address the issue of data intensity. 
In \citep{battaglia2018relational}, the relational inductive bias is widely discussed when utilizing deep learning models to deal with structured data. In recent years, \textsc{Gn} has been used to simulate physical systems \citep{tang2020towards, sanchez2020learning}, predict the structure in glassy systems \citep{bapst2020unveiling}, and simulate the molecular dynamic systems \citep{xie2019graph}. All these studies demonstrate the great potential of \textsc{Gn} in modeling the localized relationship that is viewed as the relational inductive biases. Referring to spatio-temporal networks, \textsc{Gn} is utilized to predict the climate data and the encoder-\textsc{Gn}-decoder structure is developed to strengthen the representation ability in uncovering the complex and diverse connectivity of spatio-temporal networks \citep{seo2019differentiable}. Overall, it is worth exploring to extend \textsc{Gn} to traffic prediction as the road network is also viewed as a typical spatio-temporal network

\section{Model}
\label{sec:model}
In this section, we formulate the few-sample traffic prediction task. The overall framework of \textsc{LocaleGn} is presented and each component is introduced in detail. Lastly, the computational steps in \textsc{LocaleGn} are summarized. 

\subsection{Few-sample Traffic Prediction}
The traffic prediction task can be formulated as a regression problem. Various traffic data, such as traffic speed and flow, are graph-based, which can be modeled on a spatio-temporal graph. Different types of data might associate with nodes or edges. For example, the point detector data is associated with nodes, while the travel time data is associated with edges. We define the graph associated with the traffic data as a directed graph $\mathcal{G} = (\mathcal{V},\mathcal{E})$, where each node is represented by $i$, and the set of node indices $\mathcal{V} = \{ 1, 2, \cdots ,i, \cdots, N_v \}$. $N_v$ is the number of traffic sensors. Similarly, the set of edges is defined as $\mathcal{E} = \{ 1, 2, \cdots ,k, \cdots, N_e \}$, where $N_e$ represents the number of edges. The connectivity among nodes and edges represents the relational inductive biases on the graph.

In this paper, we formulate the traffic prediction on time-dependent (dynamic) graphs. Suppose the set of time intervals during the study period is denoted as $T$, for each time interval $\tau$, we define the time-dependent data on graph $\mathcal{G}$ as $\mathbf{G}^\tau = (\mathbf{V}^\tau, \mathbf{E}^\tau)$, where $\mathbf{V}^\tau \in \mathbb{R}^{N_v \times M}$ and $\mathbf{E}^\tau \in \mathbb{R}^{N_e \times M}$. To be precise, $\mathbf{V}^\tau = [\mathbf{v}_1^\tau; \cdots; \mathbf{v}_i^\tau; \cdots; \mathbf{v}_{N_v}^\tau]$, and $\mathbf{v}_i^\tau  = [v^{\tau-M}_i, \cdots, v^{\tau}_i]^T \in \mathbb{R}^{1 \times M}$, where $\mathbf{v}_i^\tau$ represents the traffic data for node $i$ at time $\tau$. $M$ is the look-back window. The edge-based data $\mathbf{E}^\tau = [\mathbf{e}_1^\tau; \cdots; \mathbf{e}_k^\tau; \cdots; \mathbf{e}_{N_e}^\tau]$, and $\mathbf{e}_k^\tau = [e^{\tau-M}_k, \cdots, e^{\tau}_k]^T$, where $e^{\tau}_k$ represents the traffic data on edge $k$ at time $\tau$. We note the time-dependent data in $\mathbf{G}^\tau$ can not only include conventional traffic data, but also other graph-based datasets such as weather, road properties, and so on. In this paper, without loss of generality, we suppose the nodes carry the point detector data, while the edges include the road properties ({\em e.g.}, length, width, category).





Based on the above notations, existing traffic prediction models can be viewed as a function $\Phi$, and $\Phi(\mathbf{G}^\tau) = \mathbf{V}^{\tau + 1}$. $\Phi$ can be viewed as a dynamical system that evolves the graph-based traffic data from time $\tau$ to $\tau+1$. For each $\tau$, suppose $\mathbf{G}^\tau$ follows a certain distribution $\mathcal{D}$, then we have $\mathbf{G}^\tau \sim \mathcal{D}$. Given a sufficient large $T$, it is expected that we can use the deep learning model to approximate $\Phi$ such that the prediction error $\| \hat{\Phi}(\mathbf{G}^\tau) - \mathbf{V}^{\tau + 1} \|$ is minimized, where $\hat{\Phi}$ represents the deep learning-based approximator.

In the few-sample traffic prediction task, the set of archived time interval $T$ is relatively small, say $|T| = 10$, and then training $\hat{\Phi}$ can easily overfit, making the existing traffic prediction models not suitable for the few-sample tasks. To address this issue, we propose to develop a function $\Psi(\mathbf{v}_i^\tau, \mathcal{L}^\tau(i) ) = v_i^{\tau + 1}, \forall i$, where $\mathcal{L}^\tau(i)$ represents the locale of node $i$ at time $\tau$. For example, $\mathcal{L}(i)^\tau$ can represent all the $\mathbf{e}_k$ and $\mathbf{v}_{i}$ that are within the $K$-hop neighborhoods of node $i$. To obtain an approximator of $\Psi$, we notice that the available number of training data becomes $|T|N_v$. If the deep learning-based approximator $\hat{\Psi}$ could learn the \textit{locally spatial} and temporal pattern for each node with $|T|N_v$ number of samples, then the few-sample traffic prediction task can be solved. In the following sections, we present and validate \textsc{LocaleGn} as a suitable selection for $\hat{\Psi}$.

\subsection{\textsc{LocaleGn}}

\begin{figure*}[h]
  \center{\includegraphics[width=0.7\linewidth]{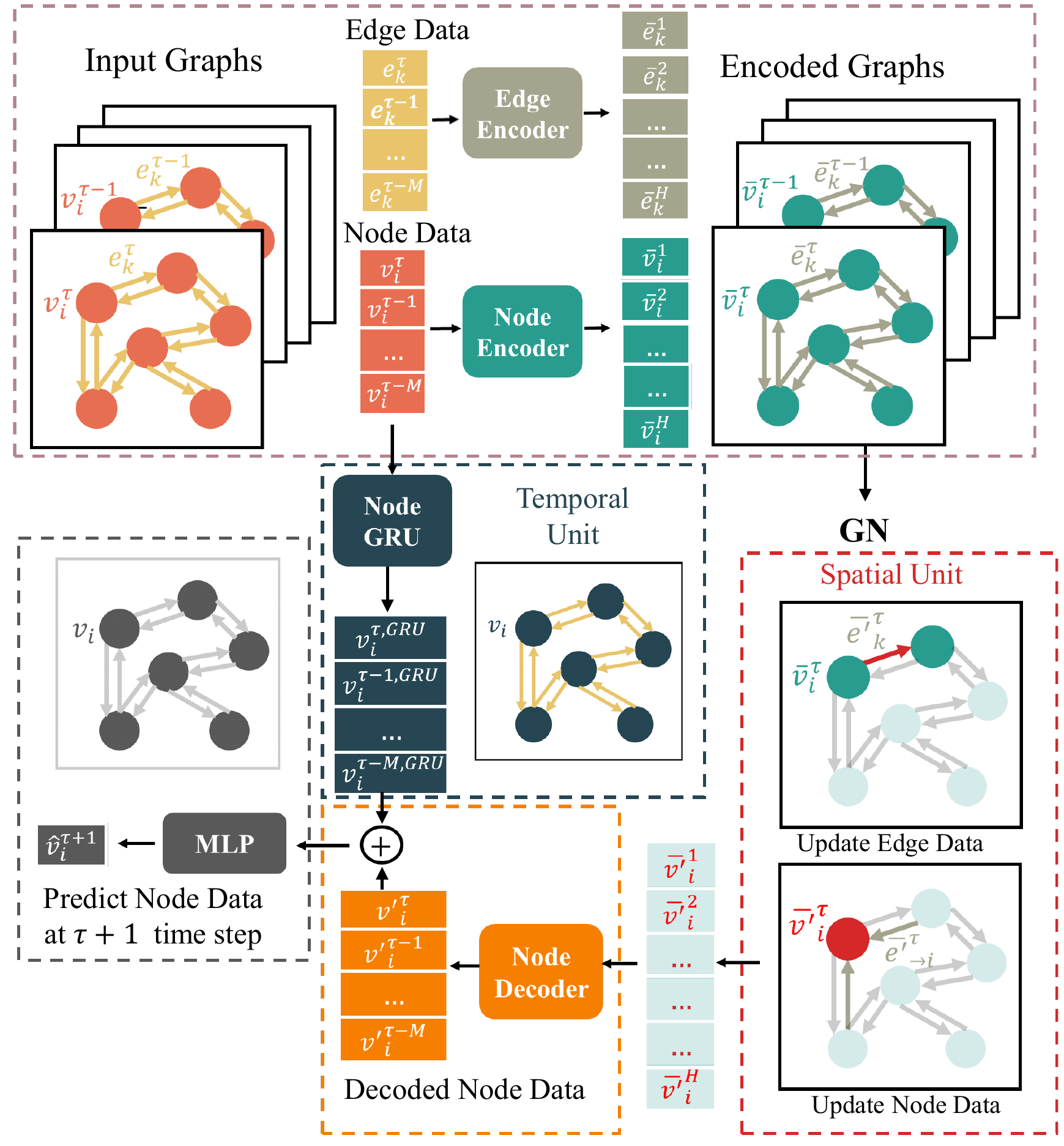}}
  \caption{An overview of \textsc{LocaleGn} ($H$ represents the size of hidden layers).}
  \label{fig:overview}
\end{figure*}

In this section, we introduce \textsc{LocaleGn} in detail. Firstly, an overview of \textsc{LocaleGn} is shown in Figure~\ref{fig:overview}.


\textsc{LocaleGn} mainly consists of four major components: node Gate Recurrent Unit (\textsc{NodeGru}), node and edge encoders, graph network (\textsc{Gn}) and nodes decoder. At each time $\tau$, the input of \textsc{LocaleGn} is $\mathbf{G}^\tau$, and the node data $\mathbf{V}^\tau$ is feed into the \textsc{NodeGru} for learning the temporal patterns of each node.  Simultaneously, both edge data $\mathbf{E}^\tau$ and node data $\mathbf{V}^\tau$ are embedded by the edge encoder and node encoder, respectively. The \textsc{Gn} is used to model the \textit{locally spatial} patterns on both nodes and edges, and the edge information is aggregated to nodes. Then, the aggregated node information is further decoded by the node decoder. Lastly, the temporal patterns by \textsc{NodeGru} is concatenated with the \textit{locally spatial} patterns by \textsc{Gn}, and a dense layer is used to generate the final prediction for $\mathbf{v}_i^{\tau+1}, \forall i$.


\subsubsection{\textsc{NodeGru}} The \textsc{NodeGru} focuses on capturing the temporal correlation on $\mathbf{v}_i$ for each node separately, as presented in Equation~\ref{eq:gru}. The temporal pattern of $\mathbf{v}_i$ is embedded in the \textsc{Gru}, which will be used for the inference of time $\tau+1$ in the following steps. We note that the \textsc{Gru} is applied for each node separately, which makes the trainable parameters dependent on the network size.   


\begin{equation}
\mathbf{v}_i^{\tau, \textsc{Gru}} = \textsc{Gru} (\mathbf{v}_i^\tau), \forall i, \tau
\label{eq:gru}
\end{equation}

\subsubsection{Node and Edge Encoder}
To encode the node and edge data in $\mathbf{G}^\tau$, we employ two multilayer perceptron $\textsc{Mlp}^{E}$ and $\textsc{Mlp}^{V}$ for edge and node respectively, as shown in Equation~\ref{eq:neen}. 

\begin{equation}
\label{eq:neen}
\begin{array}{lllll}
\overline{\mathbf{e}}_k^\tau &=& \textsc{Mlp}^{E}(\mathbf{e}_k^\tau), \forall k, \tau\\
\overline{\mathbf{v}}_i^\tau &=& \textsc{Mlp}^{V}(\mathbf{v}_i^\tau), \forall i, \tau
\end{array}
\end{equation}

The encoded node and edge data can form the encoded graph $\overline{\mathbf{G}}^\tau = (\overline{\mathbf{V}}^\tau, \overline{\mathbf{E}}^\tau)$. In general, the two encoders can better learn the latent representations of these node and edge data, and the learned representations will be used to further mine the spatial and temporal relationship by \textsc{Gn}.



\subsubsection{Graph Network}
The graph network (\textsc{Gn}) is the essential component in \textsc{LocaleGn}. In general, \textsc{Gn} models the evolution of dynamic graphs using the updating and aggregating operations on nodes and edges. In particular, we aim to use \textsc{Gn} to evolve $\overline{\mathbf{G}}^\tau$ to $\overline{\mathbf{G}}^{\tau+1}$.


In \textsc{Gn}, two updating functions $\phi^V, \phi^E$ are employed to update the per-node data and per-edge data respectively, and one aggregating function $\rho^{E \to V}$  is used to aggregate the per-edge data for each node.  To provide more details, \textsc{Gn} models the spatio-temporal propagation of the dynamic graph $\overline{\mathbf{G}}^\tau$ based on the following three steps:
\begin{enumerate}[i]
    \item In the first step, the edge updating function $\phi^E$ is applied to every single edge in the graph to calculate the per-edge updates. For each edge $k$, we combine the updated edge data  $\overline{\mathbf{e}}_k^\tau$ and the node data for both the tail node $\texttt{tail}(k)$ and head node $\texttt{head}(k)$ of $k$, and then the combined data is feed into $\phi^E$, as shown in the following equation: 
    \begin{equation}
        \overline{\mathbf{e}'}_k^{\tau} = {\phi}^{E} (\overline{\mathbf{e}}_k^\tau, \overline{\mathbf{v}}_{\texttt{tail}(k)}^\tau, \overline{\mathbf{v}}_{\texttt{head}(k)}^\tau), \forall k, \tau,
        \label{eq:edgeupdate}
    \end{equation}
    where $\phi^E$ is modeled through an MLP ($\textsc{Mlp}^{\phi^E}$).
    \item In step two, the aggregating function $\rho^{E \to V}$ is applied to all the edges that point to node $i$. Mathematically, we denote $\overline{\mathbf{E}'}_{\to i}^{\tau} = \{ \overline{\mathbf{e}'}_k^{\tau} | \texttt{tail}(k) = i, \forall k \}$ as the set of the updated edge data pointing to $i$ at time $\tau$. The function $\rho^{E \to V}$ aggregates  the information in $\overline{\mathbf{E}'}_{\to i}^{\tau}$ and  outputs a representation of all the edges pointing to node $i$, as shown in Equation~\ref{eq:agg}. We note that the aggregating function should work on different size of $\overline{\mathbf{E}'}_{\to i}^{\tau}$, and hence the element-wise mean function is chosen as $\rho^{E \to V}$. 
    \begin{equation}
    \overline{\mathbf{e}'}_{\to i}^{\tau} = {\rho}^{E \to V} (\overline{\mathbf{E}'}_{\to i}^{\tau}), \forall i, \tau\label{eq:agg}
    \end{equation}
    \item In step three, the node updating function $\phi^V$ is applied to every single node in the graph. For node $i$, the function $\phi^V$ takes the current node data $\overline{\mathbf{v}}_i^\tau$ and the representation of the edge data pointing to node $i$ computed in the previous step as input, and the output is the updated node data $\overline{\mathbf{v}'}_i^\tau$, as represented in the following equation: 
    \begin{equation}
        \overline{\mathbf{v}'}_i^\tau = {\phi}^{V} (\overline{\mathbf{e}'}_{\to i}^{\tau}, \overline{\mathbf{v}}_i^\tau), \forall i, \tau,
        \label{eq:nodeupdate}
    \end{equation}
    where $\phi^V$ is modeled as a different MLP ($\textsc{Mlp}^{\phi^V}$).
\end{enumerate}

Overall, the \textsc{Gn} model decomposes the complex topological graph structure into updating and aggregating operations on each single node and edge, and the localized relationship on the graph can be modeled accordingly. \textsc{Gn} is a generalized module that can be reduced to many existing \textsc{Gnns}, such as graph convolution neural networks, etc. Mathematically, the locale of node $i$ can be rigorously defined as $\overline{\mathbf{e}'}_{\to i}^{\tau}, \overline{\mathbf{v}}_i^\tau, \overline{\mathbf{E}'}_{\to i}^{\tau}$ and all the $\overline{\mathbf{e}}_k^\tau, \overline{\mathbf{v}}_{\texttt{tail}(k)}^\tau, \overline{\mathbf{v}}_{\texttt{head}(k)}^\tau$ for all the $k$ that are connecting to $i$. The aggregating function $\rho^{E \to V}$ can be applied to different numbers of edges for each node, making our model flexible for various graph typologies. Indeed, the updating and aggregating operations in \textsc{Gn} mimic the localized traffic exchanges shown in Figure~\ref{fig:local}, and hence it is powerful in uncovering the \textit{locally spatial} patterns of traffic data. Additionally, \textsc{Gn} is applied to each node and edge separately, so it is independent of network size, which is another attractive feature of \textsc{Gn}.

\subsubsection{Node Decoder}
The updated node $\overline{\mathbf{v}'}_i^\tau$ from the \textsc{Gn} model is further decoded by the node decoder. Similar to the node encoder, the node decoder is modeled through an MLP, as shown in Equation~\ref{eq:nodede}.

\begin{equation}
\label{eq:nodede}
\mathbf{v}_i^{'\tau} = \textsc{Mlp}^{V'}\left(\overline{\mathbf{v}'}_i^\tau\right), \forall i, \tau
\end{equation}
where $\mathbf{v}_i^{'\tau}$ represents the decoded data for node $i$ at time $\tau$.

\subsubsection{Output Layer}
In the output layer, we combine the \textit{locally spatial} and temporal information 
obtained from the \textsc{NodeGru} and node decoder, respectively. Mathematically, we concatenate $\mathbf{v}_i^{\tau, \textsc{Gru}}$ and $\mathbf{v}_i^{'\tau}$, the resulting vector is feed into an MLP for predicting the traffic states at time $\tau+1$, as shown in Equation~\ref{eq:output}. 

\begin{equation}
\label{eq:output}
\hat{v}_i^{\tau+1} = \textsc{Mlp}^{\text{Output}}\left(\mathbf{v}_i^{'\tau} \oplus \mathbf{v}_i^{\tau, \textsc{Gru}}\right), \forall i, \tau
\end{equation}
where $\oplus$ is the vector concatenation operator, $\hat{v}_i^{\tau+1}$ is the prediction for $v_i^{\tau+1}$, and $\textsc{Mlp}^{\text{Output}}$ represents another MLP. To summarize, all the computational steps in the \textsc{LocaleGn} are presented in Algorithm~\ref{alg:locale}.

Finally, the difference between the prediction $\hat{v}_i^{\tau+1}$ and the actual data $v_i^{\tau+1}$ is measured by the $\ell_2$ norm represented by $\sum_i ( v_i^{\tau+1} - \hat{v}_i^{\tau+1})^2$. The error is back-propagated to update all the parameters in \textsc{LocaleGn}.

\begin{algorithm}[h]
\caption{Computational steps in \textsc{LocaleGn}.}
\begin{algorithmic}
\REQUIRE graph structure $\mathcal{G} = (\mathcal{V},\mathcal{E})$, node data $\mathbf{V}^\tau = [\mathbf{v}_1^\tau; \cdots; \mathbf{v}_i^\tau; \cdots; \mathbf{v}_{N_v}^\tau]$ and edge data $\mathbf{E}^\tau = [\mathbf{e}_1^\tau; \cdots; \mathbf{e}_k^\tau; \cdots; \mathbf{e}_{N_e}^\tau]$.
\ENSURE \textsc{Gru} in Equation~\ref{eq:gru}, $\textsc{Mlp}^{E}$ and $\textsc{Mlp}^{V}$ in Equation~\ref{eq:neen}, $\textsc{Mlp}^{\phi^E}$ and $\textsc{Mlp}^{\phi^V}$ in \textsc{Gn}, $\textsc{Mlp}^{V'}$ in Equation~\ref{eq:nodede}, and $\textsc{Mlp}^{\text{Output}}$ in Equation~\ref{eq:output} are properly initialized and trained.\\


\FOR {$i \in \{1, \cdots, N^v \}$}
\STATE Compute $\mathbf{v}_i^{\tau, \textsc{Gru}}$ based on Equation~\ref{eq:gru}.
\STATE Compute $\overline{\mathbf{v}}_i^\tau$ based on Equation~\ref{eq:neen}.
\ENDFOR
\FOR {$k \in \{1, \dots, N^e \}$}
 \STATE Compute $\overline{\mathbf{e}}_k^\tau$ based on Equation~\ref{eq:neen}.
\STATE Update the edge data using the edge updating function based on Equation~\ref{eq:edgeupdate}.
\ENDFOR
\FOR {$i \in \{1, \cdots, N^v \}$}
\STATE Form the set of the updated edges pointing to node $i$ as $\overline{\mathbf{E}'}_{\to i}^{\tau}$.
\STATE Compute $\overline{\mathbf{e}'}_{\to i}^{\tau}$ based on Equation~\ref{eq:agg} and $\overline{\mathbf{E}'}_{\to i}^{\tau}$.
\STATE Compute $ \overline{\mathbf{v}'}_i^\tau$ based on Equation~\ref{eq:nodeupdate}.
\STATE Decode the updated node data based on Equation~\ref{eq:nodede}.
\STATE Concatenate $\mathbf{v}_i^{'\tau}$ and $\mathbf{v}_i^{\tau, \textsc{Gru}}$.
\STATE Predict $\hat{v}_i^{\tau+1}$ based on Equation~\ref{eq:output}. 
\ENDFOR
\RETURN $\hat{v}_i^{\tau+1}, \forall i \in \{1, \cdots, N^v \}$
\end{algorithmic}
\label{alg:locale}
\end{algorithm}

In addition, the computational complexity of proposed \textsc{LocaleGn} is $O(N*P +N*M)$, in which $N$ is the number of nodes, $P$ is the layer number of \textsc{Gn} and $M$ is the length of the lookback window.

\subsection{Transferability of \textsc{LocaleGn}}

In this section, we discuss how \textsc{LocaleGn} is different from the existing traffic prediction models in terms of transferability. The most important feature of \textsc{LocaleGn} is that both \textsc{Gn} and \textsc{NodeGru} modules make use of the parameter-sharing characteristic so that the knowledge learned for traffic prediction can be transferred among nodes. Though most of the existing \textsc{Gnns} can be cast into the message passing neural networks and hence the parameters can be shared \citep{gilmer2017neural}, existing traffic prediction models avoid using the parameter sharing characteristic because it could impede the prediction performance. 

In this study, we particularly consider the edge data $\mathbf{e}_t^\tau$, which represents road properties ({\em e.g.}, road length, width, and types), location information ({\em e.g.}, weather, socio-demographic factors), and so on. For node $i$, we define the locale of node $i$ at time $\tau$ to be $\mathcal{L}^\tau(i)=  \{ \mathbf{v}_i^\tau,  \overline{\mathbf{E}'}_{\to i}^{\tau}\}$, and hence the information contained in the locale can be used to conduct traffic prediction independently. The learning of traffic states becomes inductive, {\em i.e.}, \textsc{LocaleGn} universally learns the traffic states conditional on the locale, instead of learning the traffic states for each node separately. Later in the experimental experiments, we demonstrate that with the consideration of edge data, the node embedding can be learned through a universal encoder, {\em i.e.}, the \textsc{Gn} module, and the traffic prediction can be conducted per node. Additionally, the information contained in the locale is permutation-invariant, hence \textsc{LocaleGn} is suitable for different network topologies for traffic prediction.




%

\subsection{Determining the number of \textsc{Gn} layers}

One can see that the \textsc{Gn} (spatial unit) in  \textsc{LocaleGn} depicts the traffic states on each node, and the $K$ layers of \textsc{Gn} can be stacked to model the localized traffic exchanges for the $K$-hop neighbors. In this section, we discuss how to determine $K$ with theoretical analysis.

\begin{figure}[h]
  \center{\includegraphics[width=0.8\linewidth]{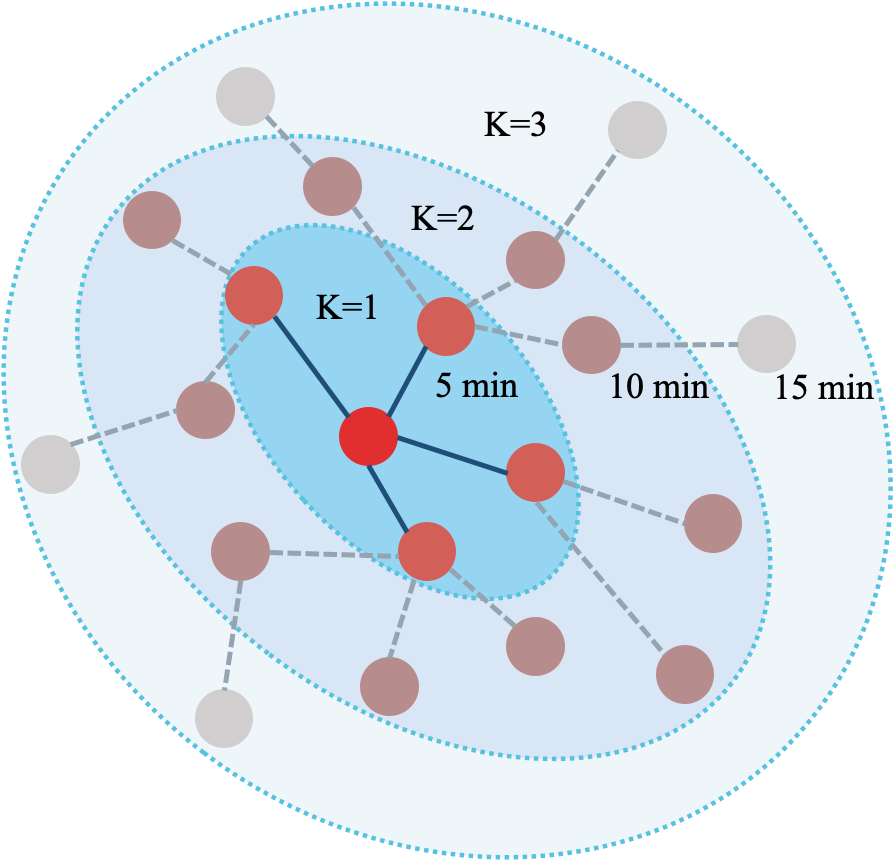}}
  \caption{Determining the number of \textsc{Gn} layers.}
  \label{fig:K_layers}
\end{figure}

 In general, traffic states are formed through the traffic flow forward propagation and congestion spillover. To predict traffic states on node $i$ for the next $L$ time intervals, we should ensure that the traffic exchanges in the next $L$ time intervals are within the $K$-hop neighbors, so that \textsc{Gn} has the capability to capture the changes in traffic states. Without loss of generality, we assume that around node $i$, the free flow speed is $f$ and the shockwave speed is $w$, respectively. Free-flow speed $f$ can be obtained from the OpenStreetMap. The OpenStreetMap contains the road level information ({\em e.g.}, highway, local roads), which can be used to infer the free-flow speed. Similarly, the shockwave speed $w$ can be determined based on the road properties in OpenStreetMap.
We define $d_i(j), \forall j \in \mathcal{V}$, which represents the distance from node $i$ to node $j$, and $K$ can be determined using Equation~\ref{eq:K}.
\begin{equation}
\label{eq:K}
\begin{array}{cccc}
K &=& \max \cup_{i \in \mathcal{V}} \left[ \mathcal{I}^{\texttt{forward}}_i \cup  \mathcal{I}^{\texttt{backward}}_i \right]\\
\mathcal{I}^{\texttt{forward}}_i &=&\left\{  \text{hop}_i(j) | d_i(j) \leq Lv,  \forall j \in \mathcal{V} \right\}\\
\mathcal{I}^{\texttt{backward}}_i &=& \left\{  \text{hop}_j(i) | d_j(i) \leq Lw, \forall j \in \mathcal{V}  \right\}
\end{array}
\end{equation}
where $\text{hop}_i(j)$ counts the number of edges in the shortest path from $i$ to $j$. $\mathcal{I}^{\texttt{forward}}_i$ and $\mathcal{I}^{\texttt{backward}}_i$ denote the sets of edge numbers to which the forward traffic flow and congestion spillover could reach from node $i$, respectively.  Using Figure~\ref{fig:K_layers} as an example, the impact of a source node can  spread to its directly connected neighbors in 5 minutes. It will take longer time to spread the impact further and hence it suffices to consider a small number of $K$ based on Equation~\ref{eq:K}.


\section{Numerical Experiments}
\label{sec:exp}
In this section, the proposed model is examined in real-world flow and speed datasets to test the short-term prediction performance.

\subsection{Datasets}
We evaluate \textsc{LocaleGn} and other baseline models on three traffic speed and three traffic flow datasets, respectively. The detailed dataset information is listed as follows:

\subsubsection{Traffic Speed Data} Three speed datasets are used and the time interval is every five minutes:
\begin{itemize}
    \item \texttt{LA}: The \texttt{LA} dataset is collected from 207 loop detectors in Los Angeles, and the data ranges from March 1st to March 7th, 2012.
    \item \texttt{SacS}: also known as PEMSD7, which contains speed data from 228 detectors in the Sacramento area. The data ranges from May to June 2012. 
    \item \texttt{ST}: \texttt{ST} contains speed data from 170 sensors in the Seattle Area. The data ranges from Jan 1st to Feb 1st, 2015. 
\end{itemize}

\subsubsection{Traffic Flow Data} Three traffic flow datasets in \citep{song2020spatial} are utilized and the time interval is every five minutes:
\begin{itemize}
    \item \texttt{SF}: also known as PEMS04, which includes 307 sensors in the San Francisco Bay area, and the data ranges from September 1st to November 7th, 2018. 
    \item \texttt{SacF}: also known as PEMS07, which contains traffic flow data from 883 detectors in the Sacramento area. The data ranges from January 1st to August 7st, 2018. 
    \item \texttt{SanB}: also known as PEMS08, which contains traffic flow data from 170 sensors in the San Bernardino area, and the data ranges from July 1st to August 7st, 2016. 
\end{itemize}

\subsection{Baseline models}
The following baseline models will be compared with \textsc{LocaleGn}. Particularly, \textsc{NodeGru} and \textsc{Gn} are presented for ablation study.
\begin{itemize}
    \item \textsc{Gcn}: Graph Convolutional Network for graph-based data prediction \citep{bruna2014spectral}.
    \item \textsc{Lstm}: Long Short-Term Memory network for time series prediction \citep{hochreiter1997long}.
    \item \textsc{T-Gcn}: Temporal Graph Convolutional Network \citep{zhao2019t}. Spatial dependency is captured by the \textsc{Gcn} module and temporal correlation is abstracted by \textsc{Gru}.
    \item \textsc{St-Gcn}: Spatial-Temporal Graph Convolutional Network \citep{yu2018spatio}. ChebNet and 2D convolutional network are utilized to capture the spatial and temporal correlation, respectively. 
    \item \textsc{Dcrnn}: Diffusion Convolutional Recurrent Neural Network \citep{li2018diffusion}.
    \item \textsc{Graph WaveNet}: Graph WaveNet for Deep Spatial-Temporal Graph Modeling \citep{wu2019graph}.
    \item \textsc{StemGNN}: Spectral Temporal Graph Neural Network \mbox{\citep{cao2020spectral}}.
    \item \textsc{Lsdlm}: Large-scale Dynamic Linear Model \mbox{\citep{kwak2021traffic}}.
    \item \textsc{NodeGru}: Gated Recurrent Unit \citep{cho-etal-2014-learning} for every single node's data in graph. This model can be viewed as \textsc{LocaleGn} with only \textsc{Gru} component.
    \item \textsc{Gn}: Graph network model with relational inductive bias \citep{li2018diffusion}. This model can be viewed as \textsc{LocaleGn} without the \textsc{NodeGru} module. 
\end{itemize}
\subsection{Experimental Settings}
\label{sec:setting}
All experiments are conducted on a desktop with Intel Core i9-10900K CPU @3.7GHz $\times$ 10, 2666MHz $\times$ 2 $\times$ 16GB RAM, GeForce RTX 2080 Ti $\times$ 2, 500GB SSD.
We divide all speed and flow datasets of seven days with a ratio of $6:1:1$ into training set, validation set and testing set. Since the most-commonly used historical data ranges from 7 days to 30 days, we choose 7 days' data as the minimum sufficient data for training the majority of existing prediction models. To simulate the situation of limited training samples, we randomly select $20\%$ of the training set for training. The sufficient dataset for prediction is 7 days (including five days' lengthen of training, 1-day lengthen of validation and 1-day lengthen of testing) and we utilize 1-day data as the insufficient training data (20\% of the original training dataset) and keep the validation and testing unchanged. When constructing $\mathbf{G}^\tau$, we set the edge data to be the normalized distance of the edges between two nodes, and the node data are the traffic speed or traffic flow from different datasets. After parameter fine-tuning with the validation set, we select one hour historical data to predict the next 5, 15, 30, 45, and 60 minutes' data, meaning $M=12$ and $L=1,3,6, 9, 12$. Hyper-parameters in \textsc{LocaleGn} are determined by the validation set, and the finalized model specifications are presented in Appendix~\ref{ap:specs}.

\subsection{Evaluation Metrics}
Three different metrics are chosen to evaluate the prediction performance of \textsc{LocaleGn} and other baseline models by comparing  $\hat{v}_i^{\tau+1}$ with the ground truth $v_i^{\tau+1}$.
\begin{itemize}
    \item Root Mean Squared Error (RMSE): \\ $\text{RMSE} = \sqrt{\frac{1}{N^v}\Sigma_{i=1}^{N^v}{(\hat{v}_i^{\tau+1} -v_i^{\tau+1})^2}}$

\item Mean Absolute Error (MAE): \\ $\text{MAE}= \frac{1}{N^v}\sum_{i=1}^{N^v}\left | \hat{v}_i^{\tau+1} -v_i^{\tau+1} \right |$
\item Mean Absolute Percentage Error (MAPE): \\$\text{MAPE} = \frac{100\%}{N^v}\sum_{i=1}^{N^v}\left| \frac{\hat{v}_i^{\tau+1} -v_i^{\tau+1}}{v_i^{\tau+1}}\right|$
\end{itemize}


\subsection{Experimental Results}
Table~\ref{table:speed} and Table~\ref{table:flow} present the prediction accuracy of different models with few training samples on three speed and flow datasets, respectively. One can see the proposed \textsc{LocaleGn} outperforms other baseline models on the six datasets for most time. \textsc{LocaleGn} outperforms the large-scale dynamic linear model-\textsc{LocaleGn} in terms of testing accuracy, which show that the knowledge learned in our proposed model can be applied to unseen scenarios.
\textsc{Lstm} utilizes temporal correlations, and \textsc{Gcn} models the spatial dependencies; both models cannot fully take the spatio-temporal dependencies into consideration at the same time. With few training samples, the performance of the state-of-the-art models like \textsc{T-Gcn} and  \textsc{St-Gcn} degrades, while \textsc{LocaleGn} achieves higher prediction accuracy. Besides, \textsc{StemGNN} achieves acceptable prediction performance but is not as accurate as \textsc{LocaleGn}
Among parameters-sharing models, \textsc{LocaleGn} also outperforms \textsc{Dcrnn} and \textsc{Graph WaveNet}. It is probably due to the complicated encoder and decoder structure of \textsc{Dcrnn} and the specified node embedding of \textsc{Graph WaveNet}.
The complicated structures make them overfit the limited available data. 
However, with parameters-sharing characteristics, \textsc{LocaleGn} is a light-weighted model with the well-designed structure that perfectly meets the requirements of short-term traffic prediction with few samples. 

We note for the flow datasets, the standard deviation is high for the baseline models, which suggests that these models may overfit with few training samples.
Besides, there is a large gap between the prediction accuracy of \textsc{NodeGru} and \textsc{Lstm}, although \textsc{Gru} and \textsc{Lstm} are similar RNN modules. 
It is because the complicated model is inclined to overfit on small datasets. 
\textsc{Gru} is less complicated than \textsc{Lstm} and \textsc{Lstm} is applied for all nodes' data while  \textsc{NodeGru} is for single node separately, which means \textsc{Lstm} have more trainable parameters and more complicated. The MAPE for some models ({\em e.g.}, \textsc{Gcn}, \textsc{Lstm}) in the flow datasets is very high, which is because the flow can be close to zero, resulting in an arbitrarily large MAPE if the flow prediction is not accurate.

\begin{table*}[h]
    \centering
    \caption{Performance of \textsc{LocaleGn} and other baseline models on traffic speed datasets  (average  $\pm$  standard deviation across 5 experimental repeats; unit for RMSE and MAE: miles/hour). }
    \footnotesize
    \resizebox{1.02\textwidth}{!}{
      \begin{tabular}{ccccccccccccc}
        \hline
        \multirow{2}*{Data} & \multirow{2}*{T} &
        \multicolumn{3}{c}{\texttt{LA}} & \multicolumn{3}{c}{\texttt{SacS}} &
        \multicolumn{3}{c}{\texttt{ST}} &
        \\
        \cmidrule(lr){3-5} \cmidrule(lr){6-8} \cmidrule(lr){9-11}
        &&RMSE  &   MAPE ($\%$)  &	MAE  &	RMSE   &	MAPE ($\%$)  &	MAE  &	RMSE   &	MAPE ($\%$)  &	MAE\\
        \hline
         \multirow{5}*{\textsc{Lsdlm}}	
        
    & 5min     & 4.81  &  6.96  & 3.39  & 2.43  & 2.61  & \bfseries 1.48  & 5.05  & 8.26  & 3.68 \\
    & 15min     & 5.96  & 8.43  & 4.09  & 3.93  & 4.94  & 2.21  & 6.01  & 10.05 & 4.23 \\
    & 30min     & 6.54  & 9.52  & 4.41  & 5.08  & 5.40   & 3.73  & 6.37  & 10.73 & 4.42 \\
   
    &45min  & 6.79  & 10.04 & 4.57  & \bfseries 5.66  & \bfseries 6.19  &  4.02  & 6.54  & 11.06 & 4.49 \\
    &60min   & 6.98  & 10.39 & 4.71  & \bfseries 6.02  & \bfseries 6.79  & 5.21  & 6.68  & 11.47 & 4.56 \\
     \arrayrulecolor{black!30}\midrule
        \multirow{5}*{\textsc{Gcn}} & 5min& 11.99  $\pm$ 	1.02	&	28.77  $\pm$ 	3.44	& 	8.55  $\pm$ 	0.65  &   11.68	 $\pm$ 	0.05	&  24.03 $\pm$ 	0.17	& 	7.96  $\pm$ 	0.15 &   9.54  $\pm$ 	0.10 	& 	19.93  $\pm$ 	0.21		& 	6.99  $\pm$ 	0.12 \\
         
        & 15min & 8.75 $\pm$ 0.02 & 16.91 $\pm$ 0.09 & 5.97 $\pm$ 0.04 & 4.47 $\pm$ 0.54 & 6.22 $\pm$ 1.05 & 2.48 $\pm$ 0.31 & 5.46 $\pm$ 0.02 & 8.76 $\pm$ 0.03 & 3.68 $\pm$ 0.04 \\
         & 30min & 9.69 $\pm$ 0.66 & 20.21 $\pm$ 2.63 & 6.84 $\pm$ 0.81 & 5.68 $\pm$ 0.07 & 7.91 $\pm$ 0.22 & 3.07 $\pm$ 0.05 & 13.11 $\pm$ 12.2 & 17.84 $\pm$ 13.76 & 8.54 $\pm$ 7.86 \\
         & 45min & 9.78 $\pm$ 0.11 & 20.19 $\pm$ 0.24 & 6.73 $\pm$ 0.11 & 6.82 $\pm$ 0.04 & 9.92 $\pm$ 0.19 & 3.74 $\pm$ 0.02 & 6.61 $\pm$ 0.04 & 10.98 $\pm$ 0.09 & 4.29 $\pm$ 0.03 \\
         & 60min & 10.20 $\pm$ 0.10 & 21.37 $\pm$ 0.38 & 7.02 $\pm$ 0.09 & 15.05 $\pm$ 6.86 & 17.33 $\pm$ 5.65 & 7.88 $\pm$ 3.52 & 9.83 $\pm$ 4.63 & 14.46 $\pm$ 3.93 & 5.85 $\pm$ 2.07 \\
    \arrayrulecolor{black!30}\midrule
          
        \multirow{5}*{\textsc{Lstm}}
        & 5min & 	12.78	 $\pm$ 	0.04	& 	29.88	 $\pm$ 	0.40	& 9.34		 $\pm$ 	0.08	&11.89		 $\pm$ 	0.19& 23.12		 $\pm$ 	0.65& 8.28	 $\pm$ 	0.08&  	9.94	 $\pm$ 	0.06		& 20.44	 $\pm$ 	0.16		& 6.70		 $\pm$ 	0.13\\
        & 15min & 9.17 $\pm$ 0.07 & 17.81 $\pm$ 0.12 & 5.82 $\pm$ 0.07 & 9.11 $\pm$ 0.26 & 15.32 $\pm$ 0.17 & 5.57 $\pm$ 0.29 & 7.74 $\pm$ 0.10 & 13.78 $\pm$ 0.21 & 5.11 $\pm$ 0.12 \\
          & 30min & 9.55 $\pm$ 0.08 & 19.63 $\pm$ 0.31 & 6.26 $\pm$ 0.05 & 9.71 $\pm$ 0.14 & 17.79 $\pm$ 0.15 & 6.12 $\pm$ 0.07 & 7.92 $\pm$ 0.09 & 14.56 $\pm$ 0.35 & 5.55 $\pm$ 0.11 \\
          & 45min & 9.85 $\pm$ 0.16 & 21.34 $\pm$ 0.84 & 6.60 $\pm$ 0.11 & 10.39 $\pm$ 0.47 & 18.98 $\pm$ 0.51 & 6.74 $\pm$ 0.20 & 8.49 $\pm$ 0.08 & 15.84 $\pm$ 0.11 & 6.02 $\pm$ 0.20 \\
          & 60min & 10.66 $\pm$ 0.14 & 23.8 $\pm$ 0.51 & 7.30 $\pm$ 0.11 & 10.78 $\pm$ 0.48 & 20.56 $\pm$ 1.08 & 7.29 $\pm$ 0.09 & 8.59 $\pm$ 0.22 & 16.38 $\pm$ 0.47 & 6.32 $\pm$ 0.14 \\
         \arrayrulecolor{black!30}\midrule
        \multirow{5}*{\textsc{T-Gcn}}	 &5min& 	5.93	 $\pm$ 	0.72	& 10.44		 $\pm$ 	1.80		& 4.02		 $\pm$ 	0.51	 &  	4.05		 $\pm$ 	0.91	& 	5.42	 $\pm$ 	1.32	& 	2.25	 $\pm$ 	0.49
        &  	5.09		 $\pm$ 	0.49	& 8.10	 $\pm$ 	0.86		& 	3.40		 $\pm$ 	0.27	\\
        & 15min & 14.43 $\pm$ 0.03 & 35.03 $\pm$ 0.32 & 10.38 $\pm$ 0.10 & 14.68 $\pm$ 0.22 & 30.15 $\pm$ 0.75 & 10.0 $\pm$ 0.36 & 11.27 $\pm$ 0.13 & 23.52 $\pm$ 0.34 & 8.36 $\pm$ 0.14 \\
          & 30min & 14.64 $\pm$ 0.19 & 35.94 $\pm$ 0.56 & 10.56 $\pm$ 0.22 & 14.54 $\pm$ 0.09 & 29.95 $\pm$ 0.43 & 10.00 $\pm$ 0.16 & 11.35 $\pm$ 0.14 & 23.87 $\pm$ 0.56 & 8.45 $\pm$ 0.08 \\
          & 45min & 14.7 $\pm$ 0.08 & 36.35 $\pm$ 0.22 & 10.62 $\pm$ 0.13 & 14.47 $\pm$ 0.01 & 29.89 $\pm$ 0.23 & 10.05 $\pm$ 0.17 & 11.30 $\pm$ 0.09 & 23.63 $\pm$ 0.16 & 8.37 $\pm$ 0.08 \\
          & 60min & 14.75 $\pm$ 0.23 & 36.45 $\pm$ 0.64 & 10.65 $\pm$ 0.22 & 14.79 $\pm$ 0.08 & 30.80 $\pm$ 0.11 & 10.09 $\pm$ 0.10 & 11.48 $\pm$ 0.05 & 24.34 $\pm$ 0.08 & 8.51 $\pm$ 0.01 \\
          \arrayrulecolor{black!30}\midrule
        \multirow{5}*{\textsc{St-Gcn}}	
        &5min& 	5.47	 $\pm$ 	0.25	&	10.39	 $\pm$ 	1.01	&	3.46	 $\pm$ 	0.18	 & 	5.50	 $\pm$ 	0.50	&	9.54	 $\pm$ 	1.14	&	3.09	 $\pm$ 	0.27& 4.38	 $\pm$ 	0.05	&	6.75	 $\pm$ 	0.10	&	2.86	 $\pm$ 	0.03\\
        
    &15min    & 7.10 $\pm$ 0.05 & 13.26 $\pm$ 0.27 & 4.28 $\pm$ 0.04 & 6.47 $\pm$ 0.12 & 10.68 $\pm$ 0.44 & 3.81 $\pm$ 0.10 & 5.60 $\pm$ 0.04 & 8.54 $\pm$ 0.13 & 3.41 $\pm$ 0.03 \\
    &30min     & 8.69 $\pm$ 0.14 & 16.95 $\pm$ 1.00 & 5.27 $\pm$ 0.10 & 8.45 $\pm$ 0.12 & 14.25 $\pm$ 0.46 & 5.21 $\pm$ 0.14 & 6.68 $\pm$ 0.08 & 10.76 $\pm$ 0.11 & 4.11 $\pm$ 0.08 \\
    &45min     & 9.99 $\pm$ 0.21 & 21.07 $\pm$ 0.77 & 6.29 $\pm$ 0.22 & 9.58 $\pm$ 0.17 & 16.79 $\pm$ 0.53 & 6.16 $\pm$ 0.34 & 7.51 $\pm$ 0.09 & 12.50 $\pm$ 0.12 & 4.66 $\pm$ 0.08 \\
    &60min    & 10.82 $\pm$ 0.28 & 23.73 $\pm$ 1.40 & 6.75 $\pm$ 0.25 & 10.21 $\pm$ 0.28 & 18.24 $\pm$ 0.55 & 6.34 $\pm$ 0.17 & 8.14 $\pm$ 0.08 & 13.89 $\pm$ 0.31 & 5.09 $\pm$ 0.13 \\
    \arrayrulecolor{black!30}\midrule
        \multirow{5}*{\textsc{Dcrnn}}	& 5min&	5.97	 $\pm$ 	0.39	&	11.11	 $\pm$ 	0.93	&	4.45	 $\pm$ 	0.39	 & 	8.42	 $\pm$ 	0.69	&	15.38	 $\pm$ 	1.50	&	5.53	 $\pm$ 	0.69  & 5.49	 $\pm$ 	0.87	&	9.24	 $\pm$ 	1.97	&	3.92	 $\pm$ 	0.87\\ 
        
    & 15min     & 5.96 $\pm$ 0.18 & 7.95 $\pm$ 0.08 & 3.43 $\pm$ 0.06 & 3.59 $\pm$ 0.22 & 3.87 $\pm$ 0.31 & 2.10 $\pm$ 0.21 & 6.32 $\pm$ 0.36 & 8.57 $\pm$ 0.49 & 3.81 $\pm$ 0.23 \\
    & 30min     & 8.18 $\pm$ 0.05 & 10.83 $\pm$ 0.28 & 4.65 $\pm$ 0.04 & 5.17 $\pm$ 0.33 & 5.1 $\pm$ 0.23 & 2.70 $\pm$ 0.12 & 7.72 $\pm$ 0.42 & 11.02 $\pm$ 0.17 & 4.73 $\pm$ 0.16 \\
    & 45min     & 8.90 $\pm$ 0.23 & 12.24 $\pm$ 0.58 & 5.03 $\pm$ 0.18 & 5.83 $\pm$ 0.43 & 5.61 $\pm$ 0.25 & 2.96 $\pm$ 0.18 & 8.84 $\pm$ 0.55 & 11.97 $\pm$ 0.61 & 5.15 $\pm$ 0.34 \\
    & 60min    & 11.01 $\pm$ 0.93 & 14.80 $\pm$ 0.64 & 6.43 $\pm$ 0.58 & 7.5 $\pm$ 1.04 & 7.31 $\pm$ 0.61 & 3.88 $\pm$ 0.31 & 9.57 $\pm$ 0.43 & 13.58 $\pm$ 0.50 & 5.78 $\pm$ 0.32 \\
    \arrayrulecolor{black!30}\midrule
        \multirow{5}*{\textsc{Graph WaveNet}}	
        &5min& 	5.30 $\pm$ 0.34 & 6.97 $\pm$ 0.46 &	3.24 $\pm$ 0.19
	 & 	5.42 $\pm$ 0.06 & 4.97 $\pm$ 0.10 & 2.67 $\pm$ 0.07  & 5.18 $\pm$ 0.21 & 7.53 $\pm$ 0.05 & 3.31 $\pm$ 0.10 \\
         &15min     & 7.16 $\pm$ 0.25 & 9.06 $\pm$ 0.26 & 4.22 $\pm$ 0.16 & 3.69 $\pm$ 0.05 & 3.76 $\pm$ 0.04 & 2.03 $\pm$ 0.03 & 6.76 $\pm$ 0.13 & 9.17 $\pm$ 0.18 & 4.07 $\pm$ 0.08 \\
     &30min     & 8.59 $\pm$ 0.31 & 10.98 $\pm$ 0.19 & 4.93 $\pm$ 0.10 & 5.62 $\pm$ 0.45 & 5.21 $\pm$ 0.55 & 2.83 $\pm$ 0.35 & 8.45 $\pm$ 0.10 & 11.54 $\pm$ 0.27 & 5.03 $\pm$ 0.11 \\
     &45min     & 9.59 $\pm$ 0.33 & 11.93 $\pm$ 0.30 & 5.22 $\pm$ 0.21 & 6.45 $\pm$ 0.17 & 5.82 $\pm$ 0.25 & 3.13 $\pm$ 0.13 & 9.62 $\pm$ 0.10 & 12.4 $\pm$ 0.21 & 5.54 $\pm$ 0.10 \\
     &60min    & 11.65 $\pm$ 0.49 & 14.76 $\pm$ 0.44 & 6.74 $\pm$ 0.34 & 8.11 $\pm$ 0.54 & 7.58 $\pm$ 1.16 & 4.18 $\pm$ 0.73 & 10.68 $\pm$ 0.19 & 13.97 $\pm$ 0.15 & 6.20 $\pm$ 0.08 \\
     \arrayrulecolor{black!30}\midrule
    \multirow{5}*{\textsc{stemGNN}}	
    &5min    &  4.55 $\pm$ 0.58 &  6.99 $\pm$ 0.83 & 3.44 $\pm$ 0.44 & 4.70 $\pm$ 0.2 & 4.91 $\pm$ 0.02 & 3.21 $\pm$ 0.01 & 7.76 $\pm$ 4.03 & 9.93 $\pm$ 5.82 & 6.24 $\pm$ 3.71 \\
    &15min     & 6.39 $\pm$ 0.31  & 8.58 $\pm$ 0.75 & 4.97 $\pm$ 0.65 & 4.44 $\pm$ 0.57 & 4.92 $\pm$ 0.67 & 3.22 $\pm$ 0.44 & 7.43 $\pm$ 3.55 & 9.47 $\pm$ 5.01 & 5.96 $\pm$ 3.20 \\
    &30min     & 6.29 $\pm$ 1.43 & 7.97 $\pm$ 1.96 & 4.42 $\pm$ 1.18 & 4.09 $\pm$ 1.28 & 4.28 $\pm$ 1.26 & 2.78 $\pm$ 0.87 & 5.73 $\pm$ 1.54 & \bfseries 7.18 $\pm$ 2.22 & 4.51 $\pm$ 1.44 \\
    &45min     & \bfseries 6.48 $\pm$ 1.42 & \bfseries 8.32 $\pm$ 1.96 & 4.54 $\pm$ 1.22 & 6.27 $\pm$ 0.61 & 6.67 $\pm$ 0.72 & 4.07 $\pm$ 0.50 & 6.38 $\pm$ 0.03 & \bfseries 8.05 $\pm$ 0.05 & 4.76 $\pm$ 0.01  \\
    &60min    & \bfseries 6.82 $\pm$ 1.22 & \bfseries 8.90 $\pm$ 1.73 & 4.93 $\pm$ 1.05 & 6.64 $\pm$ 1.31 & 7.16 $\pm$ 1.58 & 5.40 $\pm$ 1.10 & 8.43 $\pm$ 2.01 & 10.88 $\pm$ 2.55 & 6.89 $\pm$ 1.61 \\

        \arrayrulecolor{black}\midrule
        \multirow{5}*{\textsc{Gn}}	 
        &5min &4.44		 $\pm$ 	0.10		& 	7.15		 $\pm$ 	0.28		& 	2.86		 $\pm$ 	0.19 & 	2.84		 $\pm$ 	0.12	& 	4.10		 $\pm$ 	0.13		& 	1.81	 $\pm$ 	0.08		&	4.15		 $\pm$ 	0.03		& 	6.14		 $\pm$ 	0.03		& 	2.77		 $\pm$ 	0.02	\\
        
     &15min     & 8.14 $\pm$ 2.49 & 15.68 $\pm$ 6.77 & 5.98 $\pm$ 2.48 & 6.36 $\pm$ 2.13 & 11.77 $\pm$ 5.57 & 4.33 $\pm$ 1.72 & 6.36 $\pm$ 1.73 & 11.15 $\pm$ 5.36 & 4.36 $\pm$ 1.16 \\
     &30min     & 7.63 $\pm$ 1.83 & 13.66 $\pm$ 4.63 & 5.28 $\pm$ 2.00 & 7.54 $\pm$ 1.33 & 13.89 $\pm$ 3.49 & 5.52 $\pm$ 1.35 & 6.45 $\pm$ 1.47 & 11.62 $\pm$ 4.46 & 4.30 $\pm$ 1.01 \\
     &45min     & 8.31 $\pm$ 1.08 & 15.94 $\pm$ 2.53 & 5.82 $\pm$ 1.27 & 11.25 $\pm$ 1.48 & 23.49 $\pm$ 4.01 & 8.58 $\pm$ 1.05 & 8.44 $\pm$ 2.81 & 16.66 $\pm$ 7.43 & 5.78 $\pm$ 2.12 \\
     &60min   & 10.54 $\pm$ 2.62 & 22.86 $\pm$ 9.18 & 7.63 $\pm$ 1.86 & 10.68 $\pm$ 2.98 & 20.44 $\pm$ 7.37 & 8.92 $\pm$ 3.11 & 8.19 $\pm$ 2.01 & 15.80 $\pm$ 5.47 & 5.77 $\pm$ 1.81 \\
     \arrayrulecolor{black!30}\midrule
        \multirow{5}*{\textsc{NodeGru}}  
        & 5min&	4.72	 $\pm$ 	0.10	& 	7.82	 $\pm$ 	0.36		& 	2.92	 $\pm$ 	0.14
    	&	3.48		 $\pm$ 	0.17		& 	4.87		 $\pm$ 	0.39		& 	2.16	 $\pm$ 	0.18
    	&	4.29		 $\pm$ 	0.05		& 	6.59		 $\pm$ 	0.22		& 	2.81		 $\pm$ 	0.05		\\
    	
    & 15min & 5.32 $\pm$ 0.01 & 8.21 $\pm$ 0.13 & 3.23 $\pm$ 0.02 & 8.48 $\pm$ 0.30 & 17.84 $\pm$ 0.65 & 6.18 $\pm$ 0.34 & 5.09 $\pm$ 0.18 & 7.52 $\pm$ 0.52 & 3.43 $\pm$ 0.15 \\
          & 30min & 6.3 $\pm$ 0.07 & 10.09 $\pm$ 0.15 & 3.94 $\pm$ 0.14 & 6.03 $\pm$ 0.19 & 10.11 $\pm$ 0.80 & 4.33 $\pm$ 0.53 & 5.70 $\pm$ 0.36 & 9.19 $\pm$ 0.46 & 3.91 $\pm$ 0.51 \\
          & 45min & 7.58 $\pm$ 0.73 & 14.71 $\pm$ 2.33 & 4.94 $\pm$ 0.66 & 6.67 $\pm$ 0.77 & 11.25 $\pm$ 2.30 & 4.33 $\pm$ 0.68 & 6.36 $\pm$ 0.51 & 11.14 $\pm$ 1.92 & 4.23 $\pm$ 0.38 \\
          & 60min & 8.25 $\pm$ 0.32 & 16.05 $\pm$ 1.39 & 5.28 $\pm$ 0.39 & 7.60 $\pm$ 0.55 & 13.10 $\pm$ 1.88 & 5.19 $\pm$ 0.75 & 6.81 $\pm$ 0.22 & 12.24 $\pm$ 1.19 & 4.58 $\pm$ 0.09 \\
    	\arrayrulecolor{black}\midrule
        \multirow{5}*{\textsc{LocaleGn}}
        &5min	 &	\bfseries 4.24		 $\pm$ 	0.05	& \bfseries	6.88	 $\pm$ 	0.22		& 	 \bfseries 2.70		 $\pm$ 	0.05	&	\bfseries 2.56	 $\pm$ 	0.08		& \bfseries	3.72		 $\pm$ 	0.29	& \bfseries	1.65	 $\pm$ 	0.11 &\bfseries	4.00		 $\pm$ 	0.02		& \bfseries	5.96		 $\pm$ 	0.12		&\bfseries 2.65		 $\pm$ 	0.01	\\
         &15min     & \bfseries 5.25 $\pm$ 0.03 & \bfseries 7.92 $\pm$ 0.03 & \bfseries 3.21 $\pm$ 0.04 & \bfseries 3.73 $\pm$ 0.15 & \bfseries 4.85 $\pm$ 0.22 & \bfseries 2.17 $\pm$ 0.10 & \bfseries 4.86 $\pm$ 0.05 & \bfseries 6.88 $\pm$ 0.06 & \bfseries 3.15 $\pm$ 0.03 \\
     &30min    & \bfseries 6.29 $\pm$ 0.08 & \bfseries 9.96 $\pm$ 0.14 & \bfseries 3.87 $\pm$ 0.21 & \bfseries 4.99 $\pm$ 0.07 & \bfseries 6.87 $\pm$ 0.11 & \bfseries 2.92 $\pm$ 0.06 & \bfseries 5.46 $\pm$ 0.04 &  8.11 $\pm$ 0.10 & \bfseries 3.52 $\pm$ 0.06 \\
     &45min    & 7.01 $\pm$ 0.08 &  11.9 $\pm$ 0.59 & \bfseries 4.37 $\pm$ 0.23 & 6.00 $\pm$ 0.13 &  8.55 $\pm$ 0.60 & \bfseries 3.56 $\pm$ 0.20 & \bfseries 6.17 $\pm$ 0.16 & 9.27 $\pm$ 0.13 & \bfseries 4.14 $\pm$ 0.19 \\
     &60min    & 7.70 $\pm$ 0.20 &  13.08 $\pm$ 0.78 & \bfseries 4.70 $\pm$ 0.16 &  6.85 $\pm$ 0.21 &  9.20 $\pm$ 0.64 & \bfseries 4.27 $\pm$ 0.40 & \bfseries 6.50 $\pm$ 0.07 & \bfseries 9.98 $\pm$ 0.14 & \bfseries 4.11 $\pm$ 0.06 \\
       \arrayrulecolor{black}\midrule
      \end{tabular}
      }
    \label{table:speed}
  \end{table*}

\begin{table*}[h]
    \centering
    \caption{Performance of \textsc{LocaleGn} and other baseline models on traffic flow datasets (average  $\pm$  standard deviation across 5 experimental repeats; unit for RMSE and MAE: vehicles/hour).}
    \footnotesize
    \resizebox{1.02\textwidth}{!}{
     \begin{tabular}{ccccccccccccc}
        \hline
        \multirow{2}*{Data} & \multirow{2}*{T} &
        \multicolumn{3}{c}{\texttt{SF}} & \multicolumn{3}{c}{\texttt{SacF}} &
        \multicolumn{3}{c}{\texttt{SanB}} &
        \\
        \cmidrule(lr){3-5} \cmidrule(lr){6-8} \cmidrule(lr){9-11}
        && RMSE  &   MAPE ($\%$)  &	MAE  &	RMSE   &	MAPE ($\%$)  &	MAE  &	RMSE   &	MAPE ($\%$)  &	MAE\\
        \hline
         \multirow{5}*{\textsc{Lsdlm}}
          & 5min  & 31.00    & 762.92 & 20.09 & 32.56 & 23.15 & 20.37 &  32.31 & 35.35 & 20.16  \\
          & 15min & 47.24  & NA  & 31.29   & 36.53 & 43.16 & 22.52 & 33.16 & 33.85 & 20.59 \\
          & 30min &   68.80  & NA & 46.28      & 39.69 & 59.22 & 23.87 & 48.75 & 34.26 & 28.74 \\
          & 45min &   86.76  & NA & 60.00       & \bfseries 41.79 & 63.26 & \bfseries 25.28 & 62.47 & 35.83 & 36.87 \\
          & 60min &     102.50  & NA & 73.15     & \bfseries 43.99 & 65.05 & \bfseries 26.45 & 75.37 & 38.76 & 45.26 \\

        \arrayrulecolor{black!30}\midrule
        \multirow{5}*{\textsc{Gcn}}& 5min&	74.48  $\pm$ 	57.97&	90.05  $\pm$  912.20	&	93.04  $\pm$  57.97	&	105.74  $\pm$  54.29
    	&	519.16  $\pm$  143.94	&	130.18  $\pm$  143.94	& 70.82	 $\pm$  72.28&	153.15	 $\pm$  31.56 &	89.32  $\pm$  31.56\\
    	& 15min & 151.54 $\pm$ 29.17 & 317.95 $\pm$ 66.04 & 95.86 $\pm$ 28.47 & 118.33 $\pm$ 123.64 & 103.47 $\pm$ 1.78 & 83.41 $\pm$ 91.67 & 148.01 $\pm$ 88.48 & 79.13 $\pm$ 1.80 & 103.85 $\pm$ 67.25 \\
          & 30min & 72.45 $\pm$ 54.67 & 437.72 $\pm$ 68.40 & 45.43 $\pm$ 29.84 & 144.71 $\pm$ 81.83 & 111.55 $\pm$ 2.68 & 87.80 $\pm$ 47.63 & 57.56 $\pm$ 3.20 & 76.43 $\pm$ 2.97 & 39.98 $\pm$ 2.46 \\
          & 45min & 100.44 $\pm$ 67.55 & 398.08 $\pm$ 149.53 & 64.31 $\pm$ 46.91 & 123.11 $\pm$ 61.38 & 121.07 $\pm$ 4.91 & 69.38 $\pm$ 30.53 & 121.87 $\pm$ 58.46 & 77.63 $\pm$ 2.71 & 77.63 $\pm$ 35.55 \\
          & 60min & 100.62 $\pm$ 51.54 & 469.86 $\pm$ 3.42 & 62.23 $\pm$ 29.77 & 132.13 $\pm$ 65.56 & 125.25 $\pm$ 2.41 & 77.09 $\pm$ 33.87 & 119.77 $\pm$ 53.37 & 78.90 $\pm$ 2.18 & 76.14 $\pm$ 31.06 \\
          \arrayrulecolor{black!30}\midrule
    \multirow{5}*{\textsc{Lstm}}
    & 5min& 63.19  $\pm$  1.21 &	924.73  $\pm$  45.93 & 44.79  $\pm$  1.01 &	92.73  $\pm$  0.95 & 414.45  $\pm$  14.19 & 65.87  $\pm$  1.27 & 576.38  $\pm$  0.55	&	103.77  $\pm$  1.49 & 57.66  $\pm$  0.60 \\
    & 15min & 49.29 $\pm$ 1.75 & 593.17 $\pm$ 5.38 & 34.23 $\pm$ 1.38 & 147.80 $\pm$ 144.76 & 194.82 $\pm$ 13.94 & 114.67 $\pm$ 125.02 & 48.14 $\pm$ 5.24 & 53.47 $\pm$ 6.07 & 31.87 $\pm$ 3.75 \\
          & 30min & 57.72 $\pm$ 3.03 & 588.86 $\pm$ 26.15 & 40.64 $\pm$ 2.41 & 236.30 $\pm$ 145.43 & 195.43 $\pm$ 18.73 & 193.17 $\pm$ 126.75 & 49.16 $\pm$ 4.91 & 56.03 $\pm$ 3.83 & 32.78 $\pm$ 3.76 \\
          & 45min & 57.98 $\pm$ 3.93 & 628.88 $\pm$ 4.44 & 41.24 $\pm$ 3.43 & 244.19 $\pm$ 145.86 & 202.18 $\pm$ 19.33 & 200.70 $\pm$ 128.19 & 50.90 $\pm$ 4.46 & 55.47 $\pm$ 6.18 & 34.19 $\pm$ 3.04 \\
          & 60min & 65.58 $\pm$ 1.68 & 596.36 $\pm$ 18.11 & 46.84 $\pm$ 1.55 & 316.68 $\pm$ 21.14 & 178.48 $\pm$ 5.99 & 262.78 $\pm$ 21.56 & 59.21 $\pm$ 0.19 & 63.70 $\pm$ 0.68 & 40.90 $\pm$ 0.15 \\
          \arrayrulecolor{black!30}\midrule
    \multirow{5}*{\textsc{T-Gcn}}	
    &5min &		29.58		 $\pm$ 	3.04	& 210.71		 $\pm$ 	64.90		& 	19.74	 $\pm$ 	2.65	
    	&		34.86		 $\pm$ 	2.63		& 	49.41		 $\pm$ 	15.23		&	22.51		 $\pm$ 	2.31	
    	& 	31.68		 $\pm$ 	3.34		& 	50.19		 $\pm$ 	2.66		& 	21.67		 $\pm$ 	2.69	\\

          & 15min & 169.84 $\pm$ 32.15 & 359.39 $\pm$ 20.47 & 137.06 $\pm$ 26.23  & 154.76 $\pm$ 1.38 & 598.58 $\pm$ 9.44 & 126.65 $\pm$ 1.20 & 107.36 $\pm$ 0.36 & 169.32 $\pm$ 2.64 & 85.35 $\pm$ 0.30 \\
          & 30min &176.98 $\pm$ 4.16 & 391.35 $\pm$ 21.57 & 140.39 $\pm$ 2.59 & 154.96 $\pm$ 0.60 & 601.52 $\pm$ 4.69 & 126.53 $\pm$ 0.57 & 106.07 $\pm$ 0.05 & 162.40 $\pm$ 5.31 & 84.10 $\pm$ 0.03 \\
          & 45min & 102.98 $\pm$ 64.82 & 264.65 $\pm$ 76.79 & 81.33 $\pm$ 52.07 & 153.18 $\pm$ 0.95 & 582.82 $\pm$ 8.95 & 125.09 $\pm$ 0.95 & 104.79 $\pm$ 1.30 & 148.94 $\pm$ 1.61 & 83.06 $\pm$ 1.02 \\
          & 60min & 182.13 $\pm$ 0.89 & 783.17 $\pm$ 3.50 & 138.55 $\pm$ 1.92 & 152.46 $\pm$ 0.67 & 581.56 $\pm$ 3.80 & 124.14 $\pm$ 0.78 & 105.39 $\pm$ 0.49 & 142.13 $\pm$ 1.89 & 83.44 $\pm$ 0.38 \\
          \arrayrulecolor{black!30}\midrule
        \multirow{5}*{\textsc{St-Gcn}}
        &5min	&	34.10	 $\pm$ 	1.23	&	219.10	 $\pm$ 	11.24	&	23.19	 $\pm$ 	0.82 &	59.77	 $\pm$ 	6.78	&	86.98	 $\pm$ 	14.59	&	42.76	 $\pm$ 	5.09&	31.59	 $\pm$ 	0.74	&	42.29	 $\pm$ 	0.83	&	21.20	 $\pm$ 	0.55\\
        &15min	&35.19 $\pm$ 2.70 & 259.95 $\pm$ 22.33 & 24.19 $\pm$ 2.26 & 58.00 $\pm$ 9.37 & 81.47 $\pm$ 12.59 & 41.23 $\pm$ 7.53 & 35.13 $\pm$ 1.39 & 42.21 $\pm$ 0.30 & 23.68 $\pm$ 0.97 \\
    &30min	&42.05 $\pm$ 1.60 & 313.67 $\pm$ 18.94 & 30.15 $\pm$ 1.54 & 67.25 $\pm$ 10.91 & 123.56 $\pm$ 14.13 & 49.92 $\pm$ 8.70 & 42.19 $\pm$ 2.28 & 45.01 $\pm$ 0.64 & 28.41 $\pm$ 1.45 \\
    &45min	&46.22 $\pm$ 2.29 & 378.01 $\pm$ 24.73 & 33.98 $\pm$ 2.18 & 74.28 $\pm$ 9.34 & 118.77 $\pm$ 23.20 & 56.08 $\pm$ 8.47 & 45.66 $\pm$ 3.00 & 48.22 $\pm$ 0.77 & 31.64 $\pm$ 2.46 \\
    &60min	&52.36 $\pm$ 2.10 & 433.59 $\pm$ 22.07 & 38.96 $\pm$ 2.21 & 77.33 $\pm$ 3.35 & 145.14 $\pm$ 31.18 & 58.96 $\pm$ 3.12 & 52.96 $\pm$ 1.73 & 48.69 $\pm$ 0.52 & 36.68 $\pm$ 1.36 \\
    \arrayrulecolor{black!30}\midrule
        \multirow{5}*{\textsc{Dcrnn}}
         & 5min&30.32 $\pm$ 0.85 & 95.23 $\pm$ 9.24 & 21.95 $\pm$ 0.47 & 35.27 $\pm$ 2.89 & 80.56 $\pm$ 24.79 & 26.77 $\pm$ 3.55 & 32.62 $\pm$ 0.11 & 22.27 $\pm$ 1.52 & 20.41 $\pm$ 0.06 &  \\
     & 15min & 34.53 $\pm$ 4.46 & 69.32 $\pm$ 13.65 & 25.58 $\pm$ 3.61 & 44.94 $\pm$ 3.38 & 73.98 $\pm$ 24.68 & 35.68 $\pm$ 3.21 & 41.39 $\pm$ 1.41 & 14.42 $\pm$ 1.44 & 30.47 $\pm$ 1.21 \\
     & 30min & 37.74 $\pm$ 0.89 & 85.68 $\pm$ 7.61 & 27.28 $\pm$ 0.72 & 50.73 $\pm$ 4.39 & 80.00 $\pm$ 31.28 & 39.35 $\pm$ 3.85 & 44.97 $\pm$ 3.86 & 16.83 $\pm$ 0.83 & 33.37 $\pm$ 2.81 \\
    & 45min&36.93 $\pm$ 2.83 & 138.75 $\pm$ 9.44 & 26.52 $\pm$ 1.68 & 55.76 $\pm$ 7.20 & 62.95 $\pm$ 20.77 & 43.11 $\pm$ 5.56 & 50.27 $\pm$ 3.61 & 17.62 $\pm$ 3.18 & 37.64 $\pm$ 3.62 \\
    & 60min&44.41 $\pm$ 5.59 & 125.58 $\pm$ 9.54 & 32.45 $\pm$ 4.71 & 60.53 $\pm$ 1.24 & 66.49 $\pm$ 25.76 & 46.38 $\pm$ 2.15 & 51.88 $\pm$ 4.35 & 16.76 $\pm$ 2.02 & 37.99 $\pm$ 4.03 \\
    \arrayrulecolor{black!30}\midrule
        \multirow{5}*{\textsc{Graph WaveNet}}
        & 5min  & 30.90 $\pm$ 0.74 & 42.03 $\pm$ 4.57 & 22.08 $\pm$ 0.89 & 43.77 $\pm$ 1.27 & 27.60 $\pm$ 8.92 & 30.94 $\pm$ 1.90 & 38.5 $\pm$ 2.05 & 24.76 $\pm$ 1.28 & 27.53 $\pm$ 2.38 \\
       & 15min & 33.16 $\pm$ 3.27 & 61.98 $\pm$ 13.04 & 24.24 $\pm$ 2.31 & 45.07 $\pm$ 4.13 & 67.61 $\pm$ 25.89 & 35.86 $\pm$ 3.91 & 42.07 $\pm$ 1.51 & 13.50 $\pm$ 0.82 & 30.94 $\pm$ 1.40 \\
          & 30min & 37.38 $\pm$ 1.03 & 82.70 $\pm$ 8.99 & 27.11 $\pm$ 0.93 & 49.74 $\pm$ 4.86 & 58.48 $\pm$ 5.22 & 38.07 $\pm$ 3.90 & 44.35 $\pm$ 4.53 & 16.27 $\pm$ 0.27 & 32.95 $\pm$ 3.52 \\
          & 45min & 35.61 $\pm$ 2.77 & 134.1 $\pm$ 9.86 & 25.64 $\pm$ 1.37 & 55.24 $\pm$ 5.50 & 49.75 $\pm$ 13.87 & 42.93 $\pm$ 4.97 & 49.58 $\pm$ 2.33 & 15.72 $\pm$ 1.85 & 36.49 $\pm$ 2.10 \\
          & 60min & 43.98 $\pm$ 4.18 & 119.02 $\pm$ 4.50 & 31.58 $\pm$ 3.35 & 60.96 $\pm$ 0.07 & 52.07 $\pm$ 19.90 & 46.69 $\pm$ 1.24 & 49.84 $\pm$ 4.24 & 15.60 $\pm$ 1.65 & 36.22 $\pm$ 4.09 \\
          \arrayrulecolor{black!30}\midrule
    \multirow{5}*{\textsc{stemGNN}}
   & 5min  & 48.08 $\pm$ 3.31 & 36.05 $\pm$ 4.75 & 34.33 $\pm$ 4.30 & 90.73 $\pm$ 0.10 & 32.66 $\pm$ 0.04 & 67.49 $\pm$ 0.04 & 56.85 $\pm$ 0.01 & 21.05 $\pm$ 0.01 & 39.49 $\pm$ 0.01 \\
          & 15min & 29.18 $\pm$ 1.83 & 46.91 $\pm$ 1.21 & 20.72 $\pm$ 0.96 & 69.22 $\pm$ 0.01 & 28.73 $\pm$ 0.01 & 53.15 $\pm$ 0.01 & 63.71 $\pm$ 0.01 & 24.76 $\pm$ 0.01 & 48.46 $\pm$ 0.01 \\
          & 30min & 47.14 $\pm$ 2.27 & 43.82 $\pm$ 3.91 & 34.08 $\pm$ 2.08 & 89.78 $\pm$ 0.15 & 34.78 $\pm$ 0.05 & 72.28 $\pm$ 0.12 & 65.88 $\pm$ 0.08 & 25.33 $\pm$ 0.03 & 48.37 $\pm$ 0.08 \\
          & 45min & 58.56 $\pm$ 9.34 & 43.74 $\pm$ 6.73 & 43.89 $\pm$ 7.76 & 119.12 $\pm$ 0.01 & 43.58 $\pm$ 0.01 & 94.26 $\pm$ 0.01 & 68.52 $\pm$ 0.01 & 25.95 $\pm$ 0.01 & 51.21 $\pm$ 0.01 \\
          & 60min & 50.27 $\pm$ 5.27 & 48.73 $\pm$ 1.09 & 36.85 $\pm$ 4.78 & 108.37 $\pm$ 0.08 & 44.23 $\pm$ 0.02 & 85.84 $\pm$ 0.05 & 68.63 $\pm$ 0.01 & 26.18 $\pm$ 0.01 & 51.47 $\pm$ 0.01 \\
          \arrayrulecolor{black}\midrule
   
        \multirow{5}*{\textsc{Gn}}
        & 5min &25.61		 $\pm$ 	0.41	& 30.96	 $\pm$ 	2.77		& 17.84		 $\pm$ 	0.41	
    	& 	\bfseries 30.52		 $\pm$ 	0.42		& 14.65		 $\pm$ 	2.49		& \bfseries	19.77		 $\pm$ 	0.57
    	&		28.52		 $\pm$ 	1.65		& 	16.54		 $\pm$ 	0.94		& 	19.85	 $\pm$ 	1.54	\\
    & 15min & 54.61 $\pm$ 1.58 & 230.28 $\pm$ 9.67 & 44.56 $\pm$ 1.75 & 203.44 $\pm$ 81.62 & 130.01 $\pm$ 35.01 & 176.63 $\pm$ 77.74 & 130.40 $\pm$ 13.84 & 156.59 $\pm$ 11.59 & 105.09 $\pm$ 10.73 \\
          & 30min & 35.92 $\pm$ 2.41 & 107.03 $\pm$ 17.47 & 27.38 $\pm$ 2.23 & 176.08 $\pm$ 72.84 & 158.07 $\pm$ 22.82 & 146.03 $\pm$ 65.09 & 114.26 $\pm$ 23.95 & 163.91 $\pm$ 3.16 & 92.85 $\pm$ 19.24 \\
          & 45min & 37.34 $\pm$ 1.49 & 105.56 $\pm$ 11.46 & 28.37 $\pm$ 1.20 & 221.79 $\pm$ 57.30 & 148.31 $\pm$ 19.69 & 190.75 $\pm$ 54.65 & 89.50 $\pm$ 10.18 & 155.73 $\pm$ 12.78 & 72.19 $\pm$ 7.83 \\
          & 60min & 40.29 $\pm$ 0.45 & 118.27 $\pm$ 13.05 & 30.77 $\pm$ 0.43 & 218.72 $\pm$ 88.05 & 153.32 $\pm$ 27.45 & 188.17 $\pm$ 83.35 & 92.52 $\pm$ 6.83 & 167.64 $\pm$ 9.92 & 74.97 $\pm$ 6.41 \\
          \arrayrulecolor{black!30}\midrule
    	
        \multirow{5}*{\textsc{NodeGru}}	 &5min&	25.27	 $\pm$ 	0.66	& 	28.43		 $\pm$ 	11.10		& 	17.75		 $\pm$ 	1.00	
    	&	32.84	 $\pm$ 	2.44		& 	15.05  $\pm$ 	11.32	& 	22.57		 $\pm$ 	3.40	
    	&		28.12		 $\pm$ 	0.74		&  13.07		 $\pm$ 	0.78		& 	19.57		 $\pm$ 	0.74\\

   &15min     & 27.55 $\pm$ 0.06 & 39.49 $\pm$ 3.54 & 19.49 $\pm$ 0.06& 34.07 $\pm$ 0.19 & 16.20 $\pm$ 2.57 & 22.64 $\pm$ 0.31 & 219.22 $\pm$ 57.08 & 151.36 $\pm$ 27.44 & 188.55 $\pm$ 55.70 \\
    &30min  & 27.75 $\pm$ 0.27 & 45.14 $\pm$ 4.33 & 19.83 $\pm$ 0.39  & 39.09 $\pm$ 0.57 & 19.65 $\pm$ 5.60 & 26.82 $\pm$ 0.86 & 114.34 $\pm$ 65.19 & 139.66 $\pm$ 17.04 & 94.64 $\pm$ 55.35 \\
    &45min     & 31.41 $\pm$ 1.11 & 57.72 $\pm$ 7.21 & 23.12 $\pm$ 1.57  & 45.49 $\pm$ 3.97 & 28.48 $\pm$ 8.24 & 33.11 $\pm$ 4.93 & 88.88 $\pm$ 12.06 & 65.58 $\pm$ 4.04 & 74.43 $\pm$ 9.20 \\
    &60min    & 34.35 $\pm$ 1.17 & 71.13 $\pm$ 7.03 & 25.61 $\pm$ 1.70  & 47.47 $\pm$ 1.25 & 34.95 $\pm$ 13.13 & 34.29 $\pm$ 2.15 & 90.13 $\pm$ 11.68 & 53.01 $\pm$ 6.69 & 73.64 $\pm$ 10.68 \\
    \arrayrulecolor{black}\midrule

        \multirow{5}*{\textsc{LocaleGn}}	 &5min&	\bfseries 25.08		 $\pm$ 	0.85	 & 	\bfseries 23.12	 $\pm$ 	6.09	 & 	\bfseries 17.26	 $\pm$ 	0.07		 &   30.70		 $\pm$ 	1.04	 & 	\bfseries 10.53		 $\pm$ 	4.34	 & 	20.03		 $\pm$ 	0.74	 & \bfseries 26.74		 $\pm$ 	0.74		 & 	 \bfseries 13.86		 $\pm$ 	2.75	& \bfseries 18.43		 $\pm$ 	0.64\\
        
     & 15min & \bfseries 26.68 $\pm$ 0.46 & \bfseries 38.92 $\pm$ 5.86 & \bfseries 18.77 $\pm$ 0.42 & \bfseries 33.89 $\pm$ 0.16 & \bfseries 16.69 $\pm$ 1.87 & \bfseries 22.20 $\pm$ 0.25 & \bfseries 35.25 $\pm$ 1.91 & \bfseries 22.01 $\pm$ 0.17 & \bfseries 23.75 $\pm$ 1.35  \\
          & 30min & \bfseries 27.72 $\pm$ 0.46 & \bfseries 40.76 $\pm$ 5.83 & \bfseries 19.48 $\pm$ 0.48 & \bfseries 39.50 $\pm$ 1.63 & \bfseries 19.22 $\pm$ 0.90 & \bfseries 27.27 $\pm$ 2.04  & \bfseries 37.09 $\pm$ 4.88 & \bfseries 20.19 $\pm$ 2.70 & \bfseries 26.55 $\pm$ 4.28  \\
          & 45min & \bfseries 30.99 $\pm$ 0.61 & \bfseries 41.65 $\pm$ 5.73 & \bfseries 22.14 $\pm$ 0.68 &  43.77 $\pm$ 1.27 & \bfseries 27.60 $\pm$ 8.92 &  30.94 $\pm$ 1.90 & \bfseries 38.26 $\pm$ 1.69 & \bfseries 24.76 $\pm$ 1.28 & \bfseries 27.29 $\pm$ 1.94 \\
          & 60min & \bfseries 33.76 $\pm$ 0.50 & \bfseries 48.03 $\pm$ 5.90 & \bfseries 24.28 $\pm$ 0.55 &  48.74 $\pm$ 2.15 & \bfseries 30.52 $\pm$ 12.12 &  35.47 $\pm$ 2.42 & \bfseries 40.42 $\pm$ 0.42 & \bfseries 25.06 $\pm$ 2.96 & \bfseries 28.28 $\pm$ 0.49 \\

        \arrayrulecolor{black}\midrule
      \end{tabular}
       }
    \label{table:flow}
  \end{table*}

\subsubsection{Ablation Study} 
The ablation study for \textsc{LocaleGn} has been incorporated in Table~\ref{table:speed} and Table~\ref{table:flow} by comparing with \textsc{NodeGru} and \textsc{Gn}, as both models are the essential components of \textsc{LocaleGn}. In speed datasets, we observe that \textsc{LocaleGn} is consistently better than the other two models. In flow datasets, \textsc{NodeGru} and \textsc{Gn} occasionally outperform \textsc{LocaleGn}, which suggests that either the \textsc{Gru} component or \textsc{Gn} component might dominate for a specific dataset. Overall, \textsc{LocaleGn} achieves satisfactory performance for all datasets under different metrics.  

\subsubsection{Justifying the use of each module}
We further conduct more experiments to justify the use of \textsc{NodeGru} and \textsc{Gn}. To this end, we try to replace each module with other modules with similar functionality. For example, \textsc{NodeGru} can be replaced with a residual connection \citep{he2016deep}, and \textsc{Gn} can be replaced with a self-attention module \citep{vaswani2017attention}. We find that both \textsc{NodeGru} and \textsc{Gn} outperform their counterparts, which justifies the use of both modules. Details are presented in Appendix~\ref{ap:more}.

\subsection{Number of Model Parameters}
According to the few-sample experiment results,  although inferior to \textsc{LocaleGn}, \textsc{St-Gcn} achieves better prediction accuracy than other baseline models. We compare the number of trainable parameters of \textsc{LocaleGn} and \textsc{St-Gcn} in Table~\ref{tab:para}. For any network of arbitrary size, the number of trainable parameters in \textsc{LocaleGn} remains the same. However, for \textsc{St-Gcn}, the number of its parameters is positively correlated with the network size. For example, the flow dataset \texttt{SacF} has a large node number of 883, and the number of trainable parameters in \textsc{St-Gcn} is ten times larger than that of \textsc{LocaleGn}. However, the prediction performance of \textsc{LocaleGn} is even better than that of \textsc{St-Gcn}. It further proves that the short-term traffic prediction relies on localized information, and this can be efficiently learned from few samples with graph relational inductive biases. 

In the traditional graph model for spatio-temporal data prediction, the parameters sets are different for different nodes and it causes difficulty for the training process of large graphs. However, \textsc{LocaleGn} makes it feasible to learn across nodes. It can be seen as the transfer learning model among different nodes on a network. One node and its neighborhoods can be interpreted as a sub-graph, and \textsc{LocaleGn} is able to extract useful information from the sub-graphs. This suggests the reason why \textsc{LocaleGn} requires fewer samples for training.


\begin{table}[h]
    \centering
    \caption{Comparison of the number of trainable parameters between \textsc{St-Gcn} and \textsc{LocaleGn} (Unit: in thousand). }
      \begin{tabular}{llll}
        \hline
        \multirow{2}*{Model} &
        \multicolumn{3}{c}{Number of Trainable Parameters} 
        \\
        \cmidrule{2-4} 
        & \texttt{LA}  &   \texttt{SacS}  &	\texttt{ST} \\
        \hline
        \textsc{St-Gcn}	& 459 & 516 & 827\\
        \textsc{LocaleGn} & {\bf 346} & {\bf 346} & {\bf 346}\\
     
        \hline
        & \texttt{SF} & \texttt{SacF} & \texttt{SanB} \\
        \hline
        \textsc{St-Gcn}	& 768	& 4,493 & 371\\
        \textsc{LocaleGn}  & {\bf 346}	& {\bf 346} & {\bf 346}\\
        \hline
      \end{tabular}
    \label{tab:para}
  \end{table}
  
     

\subsection{Sensitivity Analysis}
\begin{figure}[h]
  \center{\includegraphics[width=1\linewidth]{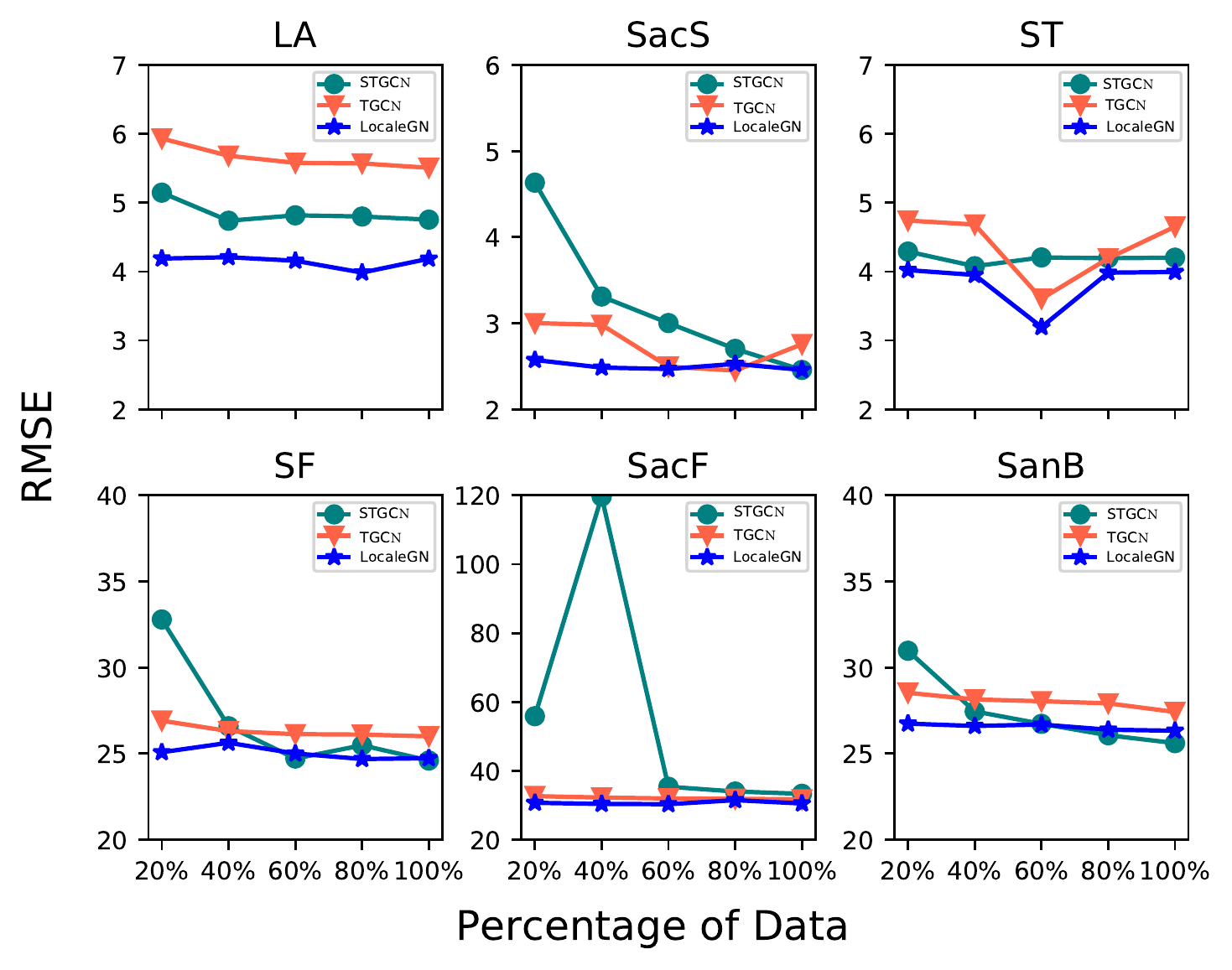}}
  \caption{Comparison of prediction performance among \textsc{LocaleGn}, \textsc{T-Gcn} and \textsc{St-Gcn} with the percentage of training set ranging from 20$\%$ to 100$\%$ (first row: speed data \texttt{LA}, \texttt{ST}, and \texttt{SacS}, unit: miles/hour; second row: flow data \texttt{SF}, \texttt{SacF}, and \texttt{SanB}, unit: vehicles/hour).}
  \label{fig:sen}
\end{figure}

Apart from evaluation in few training samples (20$\%$ of the training data), we also compare the prediction performance of \textsc{LocaleGn} and other baseline models with different percentages of training data (from 20$\%$ to 100$\%$), the results are presented in Figure~\ref{fig:sen}. The prediction performance remains stable for \textsc{LocaleGn} when the training ratio decreases from 100$\%$ to 20$\%$, while the prediction performance of \textsc{St-Gcn} gradually degrades when reducing the number of training samples.

Overall, \textsc{LocaleGn} outperforms \textsc{T-Gcn} and \textsc{St-Gcn} with all percentages of training data and the advantage is notably large when the ratio is 20$\%$ or 40$\%$. In other words, \textsc{T-Gcn} and \textsc{St-Gcn} overfit and become unstable when the training set is small. In contrast, \textsc{LocaleGn} has better learning ability in capturing the localized spatial and temporal correlations than \textsc{T-Gcn} and \textsc{St-Gcn}. The results further demonstrate that \textsc{LocaleGn} is a powerful and efficient model for the task of the few-sample traffic prediction.  

\subsection{Cross-city Transfer Analysis}

\begin{figure*}[h]
  \center{\includegraphics[width=0.8\linewidth]{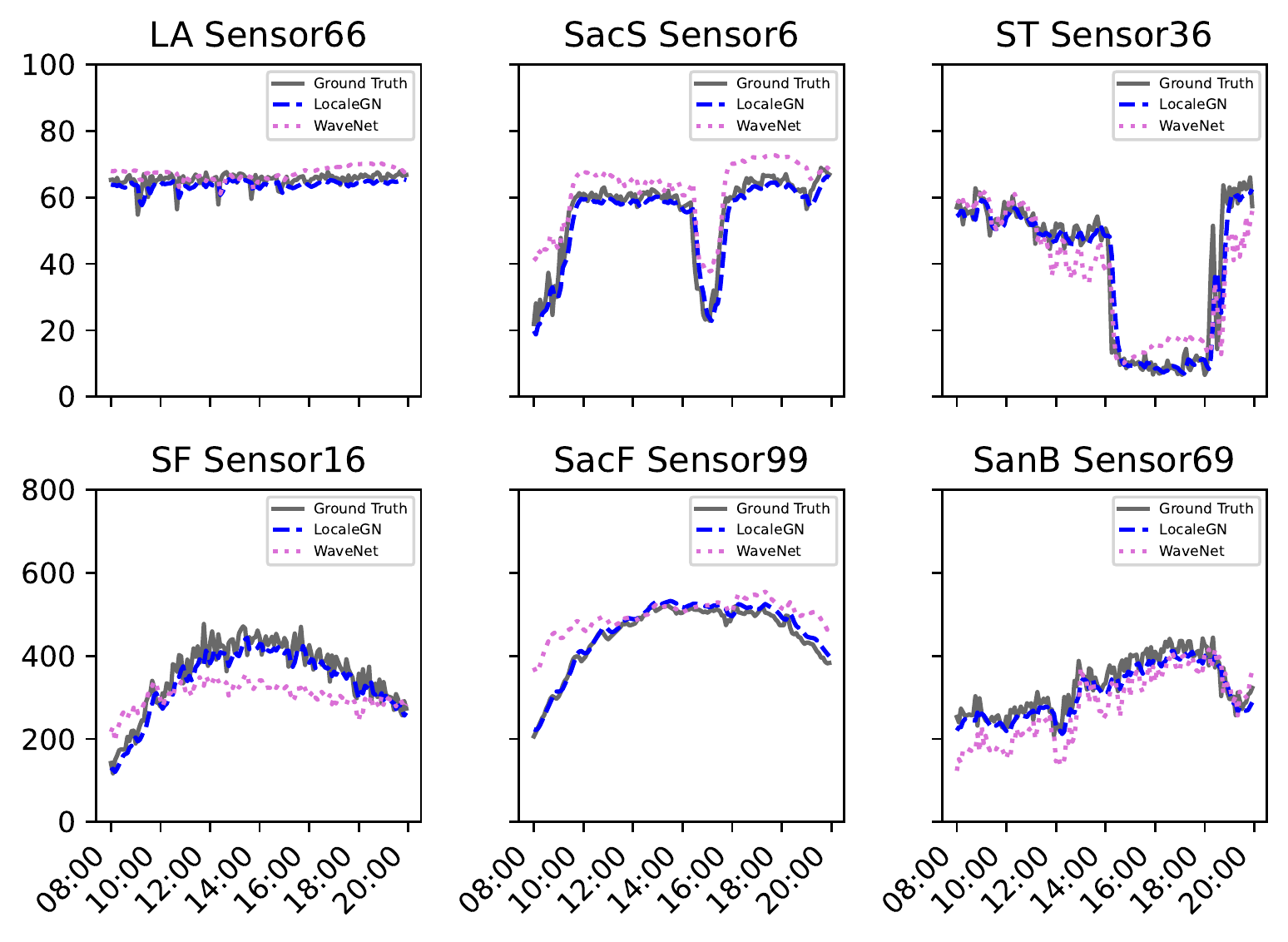}}
  \caption{Comparison of prediction performance with \textsc{LocaleGn} and \textsc{GraphWaveNet} for cross-city transfer analysis (first row: speed data \texttt{LA}, \texttt{SacS}, and \texttt{ST}, unit: miles/hour; second row: flow data \texttt{SF}, \texttt{SacF}, and \texttt{SanB}, unit: vehicles/hour).}
  \label{fig:tansfer analysis}
\end{figure*}

In this section, we consider a more challenging situation, in which traffic sensors are just installed or the traffic prediction service is just deployed in a city. In this case, we would like to use the knowledge learned from some city and directly apply it to a new city. To this end, we consider one city (target graph) that does not have historical traffic data, while some other cities (source graphs) have archived historical data. One can see this setting focuses on the transfer learning across cities.
The well-performed \textsc{LocaleGn} model is designed for extracting locally spatial and temporal traffic patterns, which is independent of graph structures and has the potential to be transferred to the target city. As the spectral graph-based \textsc{Gnn} can be applied to graph structures, we compare \textsc{LocaleGn} with \textsc{Dcrnn} and \textsc{Graph WaveNet}, the two state-of-the-art spectral graph-based models.

We compare the three models on speed and flow datasets respectively. Each model is pre-trained on a source graph with a random 20$\%$ of the training data and then directly tested on the target graph without further training. The prediction results on the target graph are presented in Table~\ref{table:transfer_speed} and Table~\ref{table:transfer_flow}.  One can see that \textsc{LocaleGn} consistently outperforms  \textsc{Dcrnn} and \textsc{Graph WaveNet} on different cities for both speed and flow datasets, and the prediction accuracy is very stable. 
The results show that \textsc{LocaleGn} has great potential to transfer pre-trained models across cities even without fine-tuning the data of the target city because the universal traffic patterns and physical rules regardless of network structures can be learned by \textsc{LocaleGn}. 

In Figure~\ref{fig:tansfer analysis}, we compare the prediction results of \textsc{Graph WaveNet} and \textsc{LocaleGn} from 8:00 AM to 20:00 PM for three traffic speed datasets and three traffic flow datasets, respectively. In general, \textsc{LocaleGn} can make more accurate predictions than \textsc{Graph WaveNet} in all datasets. 
It is notable that when the speed/flow surges or decreases sharply, the performance of \textsc{Graph WaveNet} will decline significantly while the prediction of \textsc{LocaleGn} remains stable. 
This demonstrates that although \textsc{Graph WaveNet} and \textsc{Dcrnn} contain the parameters-sharing characteristics, they cannot learn the universal rule for short-term traffic prediction and their performance still depends on graph structures. Consequently,  directly applying \textsc{Graph WaveNet} and \textsc{LocaleGn} to the target city will result in low prediction accuracy. In contrast, the performance of \textsc{LocaleGn} remains satisfactory. The structure of \textsc{LocaleGn} can serve as an essential building block for transferring prediction models between cities. Besides, the embedding and temporal components can be easily replaced with other state-of-the-art deep learning modules. 

\begin{table*}[h]
    \centering
    \caption{Transferring Performance of \textsc{Dcrnn}, \textsc{Graph WaveNet} and \textsc{LocaleGn} on traffic speed datasets  (average  $\pm$  standard deviation across 5 experimental repeats; unit for RMSE and MAE: miles/hour). }
    \footnotesize
    \resizebox{1.02\textwidth}{!}{
      \begin{tabular}{lllllllllllll}
        \hline
        \multirow{2}*{Source/Target} &
        \multicolumn{3}{c}{\textsc{Dcrnn}} & \multicolumn{3}{c}{\textsc{Graph WaveNet}} &
        \multicolumn{3}{c}{\textsc{LocaleGn}} &
        \\
        \cmidrule(lr){2-4} \cmidrule(lr){5-7} \cmidrule(lr){8-10}
        &RMSE  &   MAPE ($\%$)  &	MAE  &	RMSE   &	MAPE ($\%$)  &	MAE  &	RMSE   &	MAPE ($\%$)  &	MAE\\
        \hline
        \texttt{LA}/\texttt{SacS}	 &	4.52	 $\pm$ 	0.39	&	7.56	 $\pm$ 	0.74	&	3.30	 $\pm$ 	0.39	&	 7.34	 $\pm$ 	0.65	&	13.73	 $\pm$ 	1.48	&	5.88	 $\pm$ 	0.65 & \bfseries 3.84	 $\pm$ 	0.31		& \bfseries	5.56		 $\pm$ 	0.69	& \bfseries	2.79	 $\pm$ 	0.47 	\\
        \texttt{LA}/\texttt{ST}	 &		4.93	 $\pm$ 	0.37	&	7.87	 $\pm$ 	0.74	&	3.71	 $\pm$ 	0.37	&	6.59	 $\pm$ 	0.20	&	10.12	 $\pm$ 	0.77	&	4.89	 $\pm$ 	0.20 &	\bfseries 4.00		 $\pm$ 	0.02		& \bfseries	5.96		 $\pm$ 	0.12		& \bfseries 2.65		 $\pm$ 	0.01	\\
        \texttt{SacS}/\texttt{LA}	 &		9.90	 $\pm$ 	1.20	&	19.93	 $\pm$ 	2.46	&	6.57	 $\pm$ 	1.20	&	 7.15	 $\pm$ 	0.32	&	13.20	 $\pm$ 	1.51	&	5.06	 $\pm$ 	0.32 & \bfseries 5.07		 $\pm$ 	0.50	& \bfseries 	8.09	 $\pm$ 	1.02		& \bfseries	  3.65		 $\pm$ 	0.70	\\
        \texttt{SacS}/\texttt{ST}	 &		7.29	 $\pm$ 	0.43	&	13.44	 $\pm$ 	0.98	&	4.98	 $\pm$ 	0.43	&	 7.34	 $\pm$ 	0.82	&	10.78	 $\pm$ 	0.87	&	5.42	 $\pm$ 	0.82 & \bfseries 4.57		 $\pm$ 	0.45		& \bfseries	6.80		 $\pm$ 	0.89		& \bfseries 3.24		 $\pm$ 	0.59	\\
        \texttt{ST}/\texttt{LA}	 &		8.22	 $\pm$ 	1.33	&	15.98	 $\pm$ 	3.63	&	5.87	 $\pm$ 	1.33	&	 6.88	 $\pm$ 	0.64	&	12.94	 $\pm$ 	1.03	&	4.70	 $\pm$ 	0.64 & \bfseries 4.86		 $\pm$ 	0.21	&  \bfseries	7.62	 $\pm$ 	0.52		& 	 \bfseries 3.33		 $\pm$ 	0.35
        \\
        \texttt{ST}/\texttt{SacS}	 &		6.90	 $\pm$ 	2.18	&	12.19	 $\pm$ 	4.17	&	4.64	 $\pm$ 	2.18	&	 7.32	 $\pm$ 	1.08	&	13.06	 $\pm$ 	1.63	&	5.76	 $\pm$ 	1.08& \bfseries 3.62	 $\pm$ 	0.24		& \bfseries	5.07		 $\pm$ 	0.56	& \bfseries	2.44	 $\pm$ 	0.37
        \\
       \arrayrulecolor{black}\midrule
      \end{tabular}
      }
    \label{table:transfer_speed}
  \end{table*}

\begin{table*}[h]
    \centering
    \caption{Transferring Performance of \textsc{Dcrnn}, \textsc{Graph WaveNet} and \textsc{LocaleGn} on traffic flow datasets  (average  $\pm$  standard deviation across 5 experimental repeats; unit for RMSE and MAE: vehicles/hour). }
    \footnotesize
    \resizebox{1.02\textwidth}{!}{
      \begin{tabular}{lllllllllll}
        \hline
        \multirow{2}*{Source/Target} &
        \multicolumn{3}{c}{\textsc{Dcrnn}} & \multicolumn{3}{c}{\textsc{Graph WaveNet}} &
        \multicolumn{3}{c}{\textsc{LocaleGn}} &
        \\
        \cmidrule(lr){2-4} \cmidrule(lr){5-7} \cmidrule(lr){8-10}
        &RMSE  &   MAPE ($\%$)  &	MAE  &	RMSE   &	MAPE ($\%$)  &	MAE  &	RMSE   &	MAPE ($\%$)  &	MAE\\
        \hline
        \texttt{SF}/\texttt{SacF}	 &125.59	 $\pm$ 	9.02	&	571.77	 $\pm$ 	145.49	&	96.82	 $\pm$ 	9.02	&	129.38	 $\pm$ 	12.20	&	106.95	 $\pm$ 	27.87	&	108.96	 $\pm$ 	12.20 &	\bfseries 31.88	 $\pm$ 	0.87		& \bfseries	15.26		 $\pm$ 	4.81	& \bfseries	21.57	 $\pm$ 	1.34	\\
        \texttt{SF}/\texttt{SanB}	 &102.19	 $\pm$ 	6.23	&	299.38	 $\pm$ 	63.31	&	78.83	 $\pm$ 	6.23	&90.58	 $\pm$ 	6.23	&	103.88	 $\pm$ 	26.38	&	77.27	 $\pm$ 	6.23 &	\bfseries 27.17		 $\pm$ 	0.67		& 	\bfseries 14.00		 $\pm$ 	2.14		& \bfseries 19.00		 $\pm$ 	0.88	\\
        \texttt{SacF}/\texttt{SF}	 &	84.00	 $\pm$ 	5.75	&	659.87	 $\pm$ 	154.49	&	61.94	 $\pm$ 	5.75	&	 90.76	 $\pm$ 	4.39	&	85.02	 $\pm$ 	8.83	&	78.63	 $\pm$ 	4.39& \bfseries 25.87		 $\pm$ 	1.01	&  	\bfseries 21.71	 $\pm$ 	2.65		& \bfseries	  18.02		 $\pm$ 	1.06	\\
        \texttt{SacF}/\texttt{SanB}	 &		95.80	 $\pm$ 	6.31	&	331.29	 $\pm$ 	40.89	&	74.75	 $\pm$ 	6.31	&	 87.77	 $\pm$ 	5.51	&	90.50	 $\pm$ 	10.88	&	75.39	 $\pm$ 	5.51 & \bfseries 27.93		 $\pm$ 	1.67		& \bfseries	13.40		 $\pm$ 	0.74		& \bfseries 19.67		 $\pm$ 	1.78	\\
        \texttt{SanB}/\texttt{SF}	 &		88.32	 $\pm$ 	7.09	&	910.14	 $\pm$ 	140.04	&	63.38	 $\pm$ 	7.09&	 81.96	 $\pm$ 	11.46	&	95.34	 $\pm$ 	15.22	&	73.89	 $\pm$ 	11.46 & \bfseries 26.44		 $\pm$ 	1.76	&  \bfseries	30.95	 $\pm$ 	11.12	&	\bfseries 18.97		 $\pm$ 	2.08	\\
        \texttt{SanB}/\texttt{SacF}	 &		123.60	 $\pm$ 	10.77	&	591.62	 $\pm$ 	76.52	&	94.44	 $\pm$ 	10.77	&	 110.59	 $\pm$ 	18.08	&	108.27	 $\pm$ 	23.19	&	99.32	 $\pm$ 	18.08 & \bfseries 33.83	 $\pm$ 	3.51	& \bfseries	19.19		 $\pm$ 	8.19	& \bfseries	24.02	 $\pm$ 	4.30	\\
       \arrayrulecolor{black}\midrule
      \end{tabular}
      }
    \label{table:transfer_flow}
  \end{table*}

\section{Conclusion}
\label{sec:con}
In this paper, we discuss and define the research question of few-sample traffic prediction on large-scale networks. A graph network-based model \textsc{LocaleGn} is developed to learn the \textit{locally spatial} and temporal patterns of traffic data, thus accurate short-term predictions can be generated. 
The parameter-sharing characteristics help \textsc{LocaleGn} prevent overfitting with limited training data. Additionally, the learned knowledge in \textsc{LocaleGn} can be transferred across cities.
Extensive evaluations on six real-world traffic speed or flow datasets demonstrate that \textsc{LocaleGn} outperforms other baseline models. \textsc{LocaleGn} is also more light-weighted as it contains less trainable parameters, and this also suggests a shorter training time and lower data requirements. 
Ablation study and sensitivity analysis are conducted to show the compactness and robustness of \textsc{LocaleGn}. 
The cross-city transfer analysis also demonstrates its great potential in developing traffic prediction services without training in a new city.
Overall, this paper sheds light on utilizing the transferability of \textsc{LocaleGn} for traffic prediction and developing traffic prediction models for cities with few historically archived data.


As for the future research directions, firstly, it is necessary to interpret the knowledge learned by the \textsc{LocaleGn} among different nodes, as this paper has demonstrated that the learned knowledge among nodes can contribute to the prediction accuracy. 
Secondly, because \textsc{LocaleGn} can be applied to different cities, it is interesting to develop a unified and fair framework for training and testing the \textsc{LocaleGn} on multiple cities without biases.
Essentially, \textsc{LocaleGn} has capacities in transferring knowledge not only among nodes in a single graph, but also across different graphs. Thirdly, \textsc{LocaleGn} can be further extended to predict OD demand for public transit and ride-hailing services \citep{geng2019spatiotemporal,yao2018deep}. Indeed, the enlightening idea of relational inductive bias and the simple structure of \textsc{LocaleGn} can be applied to learn various spatio-temporal datasets, and the localized patterns can be extracted. These datasets include human mobility patterns, social media, epidemics, and climate-related data.



\appendix
\subsection{Model Specifications}
\label{ap:specs}
In this section, we present details of the model specifications of \textsc{LocaleGn}. 
There are four modules in \textsc{LocaleGn}, which include Encoder, \textsc{NodeGru}, \textsc{Gn}, and the Output layer. The detailed layer information of each module is listed in Table~\ref{tab:appendix}. 
 \begin{table}[h]
    \centering
    \caption{Functions in \textsc{LocaleGn}.}
    \footnotesize
      \begin{tabular}{cc}
        \arrayrulecolor{black}\midrule
        \textbf{Module} & \textbf{Functions}
        \\
        \hline
        \\[-1em]
        Node Encoder & DenseLayer [12,64] with ReLU  \\
        \\[-1em]
        Edge Encoder & DenseLayer [1,64] with ReLU\\
        \\[-1em]
        \hline
        \\[-1em]
        \textsc{NodeGru} &   [12,64] \\
        \\[-1em]
        \hline
        \\[-1em]
        \multirow{2}{*}{Updating by ${\phi}_{E}$}&$\overline{\mathbf{e}'}_k^{\tau} = {\phi}^{E} (\overline{\mathbf{e}}_k^\tau, \overline{\mathbf{v}}_{\texttt{tail}(k)}^\tau, \overline{\mathbf{v}}_{\texttt{head}(k)}^\tau), \forall k, \tau$\\
        \\[-1em]
        &DenseLayer [256,64] with ReLU\\
        \\[-1em]
        Aggregating by ${\rho}^{E \to V}$ & Element-wise average operator\\
        \\[-1em]
        \multirow{2}{*}{Updating by ${\phi}_{V}$} & $\overline{\mathbf{v}'}_i^\tau = {\phi}^{V} (\overline{\mathbf{e}'}_{\to i}^{\tau}, \overline{\mathbf{v}}_i^\tau), \forall i, \tau$  \\
        \\[-1em]
        & DenseLayer [256,64] with ReLU\\
        \\[-1em]
        \hline
        \\[-1em]
        Node Decoder & DenseLayer [64,64] with ReLU\\
        \\[-1em]
        \hline
        \\[-1em]
        \multirow{2}{*}{Output layer} & Concatenation operator and\\
        \\[-1em]
        & DenseLayer [64,1] with ReLU \\
        \arrayrulecolor{black}\midrule
      \end{tabular}
    \label{tab:appendix}
  \end{table}
 
Besides, we provide hyper-parameters for model setting and training process in Table~\ref{tab:appendix_2}. 
 
\begin{table}[h]
    \centering
    \caption{Hyper-parameters in \textsc{LocaleGn}.}
      \begin{tabular}{cc}
        \arrayrulecolor{black}\midrule
        {\bf Hyper-parameters} & {\bf Values}
        \\
        \hline
        \\[-1em]
        Optimizer &  Adam \\
        \\[-1em]
        Learning rate &  0.001 \\
        \\[-1em]
        Weight decay & 0.0005 \\
        \\[-1em]
        Iteration & $3,000$\\
        \\[-1em]
        Lookback window & 12 \\
        \\[-1em]
        Node hidden dimension & 64  \\
        \\[-1em]
        Edge hidden dimension & 64\\
        \\[-1em]
        Number of \textsc{Gru} layer & 1\\
        \\[-1em]
        Number of \textsc{Gn} layer& 1\\
        \arrayrulecolor{black}\midrule
      \end{tabular}
    \label{tab:appendix_2}
  \end{table}

\subsection{Additional Experiments}
\label{ap:more}
In this section, we provide results of some additional experiments to further justify the proposed structure of \textsc{LocaleGn}.

\subsubsection{Residual Graph Network}
On top of the basic architecture of Encoder, \textsc{Gn}, and Decoder, we also try to add residual connections to the \textsc{Gn} module.  Instead of using \textsc{NodeGru}, the \textsc{Gn} module is developed to learn the differences between the input graph and the expected output graph. The encoded data will be processed by the \textsc{Gn} module, in which we add the residual connection to update the node attributes. The design of  Residual Graph Network (\textsc{RGn}) model is inspired by the theory of dynamical system, in which the output at $\tau+1$ can be calculated based on the input at time $\tau$ as well as the evolution function \textsc{Gn}. Mathematically, the residual connection can be presented in Equation~\ref{eq:RGN}.
\begin{equation}
\overline{G}_{\tau+1} = \textsc{Gn} (\overline{G}_{\tau}) + \overline{G}_{\tau}
\label{eq:RGN}
\end{equation} 

\subsubsection{Attention Graph Network}
We also try to replace the \textsc{NodeGru} module in \textsc{LocaleGn} with the \textsc{Self-Attention} mechanism to capture the temporal dependency of the traffic data. Different from applying \textsc{Gru} for each node data separately, the \textsc{Self-Attention} module is applied to all the node data simultaneously. The reason we compare with the \textsc{Self-Attention} module is that recent studies indicate its strong potential in transferability \citep{vaswani2017attention, DBLP:conf/naacl/DevlinCLT19}. The proposed Attention Graph Network (\textsc{AGn}) can be represented in Equation~\ref{eq:attn}. 
\begin{equation}
\overline{G}_{\tau+1} = \textsc{Gn} (\overline{G}_{\tau}+\textsc{Self-Attention}(\overline{G}_{\tau})) 
\label{eq:attn}
\end{equation} 
where the key, query and value in \textsc{Self-Attention} are embedded with historical node data. The outcome vector from \textsc{Self-Attention} is transformed to a lower dimension vector to obtain the compressed temporal patterns, which are then combined with the original node data. The combined vector is fed into an encoder to generate the input of \textsc{Gn} in order to further infer the spatio-temporal patterns at time $\tau + 1$.

\subsubsection{Results Comparison}

Using the same experiment settings as in section~\ref{sec:setting}, the additional two models are examined on the three speed datasets and three flow datasets, respectively.
The prediction errors of \textsc{LocaleGn}, \textsc{RGn} and \textsc{AGn} are listed in Table~\ref{table:appendix_speed} and Table~\ref{table:appendix_flow}. 

\begin{table}[h]
    \centering
    \caption{Performance of \textsc{LocaleGn} , \textsc{RGn} and \textsc{AGn} on traffic speed datasets.}

      \begin{tabular}{lllll}
        \hline
        \multirow{2}*{Model} &
        \multicolumn{3}{c}{\texttt{LA}} & 
        \\
        \cmidrule(lr){2-4} 
        &RMSE  &   MAPE ($\%$)  &	MAE \\
        \hline
        \textsc{AGn} & 4.45 &   7.18	&  2.80 \\
        \textsc{RGn}	& 	4.34	& 	6.94 & 2.73	\\
    	\arrayrulecolor{black!30}\midrule
        \textsc{LocaleGn}	 &	\bfseries	4.24		&  \bfseries	6.88	& 	 \bfseries 2.70	 \\
        \arrayrulecolor{black}\midrule
        \multirow{2}*{} &
        \multicolumn{3}{c}{\texttt{SacS}} & 
        \\
        \cmidrule(lr){2-4} 
        &RMSE  &   MAPE ($\%$)  &	MAE \\
        \textsc{AGn} & 	2.94 &   4.21 	& 	1.91 \\
        \textsc{RGn} & \bfseries 2.55	& \bfseries 3.53 & \bfseries 1.58\\
        \arrayrulecolor{black!30}\midrule
        \textsc{LocaleGn} & 2.56	&	3.72 & 	1.65\\
        
        \arrayrulecolor{black}\midrule
        \multirow{2}*{} &
        \multicolumn{3}{c}{\texttt{ST}} & 
        \\
        \cmidrule(lr){2-4} 
        &RMSE  &   MAPE ($\%$)  &	MAE \\
         \textsc{AGn} & 	4.13 & 6.14 & 2.73\\
         \textsc{RGn} &  4.09	& 6.06	& 2.77\\
         \arrayrulecolor{black!30}\midrule
         \textsc{LocaleGn} &\bfseries	4.00 & \bfseries	5.96	&\bfseries 2.65\\
       \arrayrulecolor{black}\midrule
      \end{tabular}
    \label{table:appendix_speed}
  \end{table}
  
\begin{table}[h]
    \centering
    \caption{Performance of \textsc{LocaleGn} , \textsc{RGn} and \textsc{AGn} on traffic flow datasets.}
    \footnotesize
      \begin{tabular}{lllll}
        \hline
        \multirow{2}*{Model} &
        \multicolumn{3}{c}{\texttt{SF}} & 
        \\
        \cmidrule(lr){2-4} 
        &RMSE  &   MAPE ($\%$)  &	MAE \\
        \hline
        \textsc{AGn} & 	\bfseries 25.00 &	40.49 &	\bfseries 17.25 \\
        \textsc{RGn} & 25.86 &	34.42 & 18.02	\\
    	\arrayrulecolor{black!30}\midrule
        \textsc{LocaleGn}	 &	 25.08		& 	\bfseries 23.12	& 	17.26 \\
        \arrayrulecolor{black}\midrule
        \multirow{2}*{} &
        \multicolumn{3}{c}{\texttt{SacF}} & 
        \\
        \cmidrule(lr){2-4} 
        &RMSE  &   MAPE ($\%$)  &	MAE \\
        \textsc{AGn} & 30.73 & 15.58 & 20.19 \\
        \textsc{RGn}&	\bfseries 30.34 & 12.79 & \bfseries19.64 \\
        \arrayrulecolor{black!30}\midrule
        \textsc{LocaleGn} &   30.70		& 	\bfseries 10.53		 & 	20.03\\
        
        \arrayrulecolor{black}\midrule
        \multirow{2}*{} &
        \multicolumn{3}{c}{\texttt{SanB}} & 
        \\
        \cmidrule(lr){2-4} 
        &RMSE  &   MAPE ($\%$)  &	MAE \\
         \textsc{AGn} & 34.97 & 22.60 & 26.56\\
         \textsc{RGn}  & 28.23 &	19.30 & 19.77\\
         \arrayrulecolor{black!30}\midrule
         \textsc{LocaleGn} & \bfseries 26.74 &  \bfseries	 14.86		& \bfseries 18.43\\
       \arrayrulecolor{black}\midrule
      \end{tabular}
    \label{table:appendix_flow}
  \end{table}  
 
For traffic speed prediction, \textsc{LocaleGn} outperforms \textsc{RGn} and \textsc{AGn} on both \texttt{LA} and \texttt{ST}, but slightly under-performs \textsc{RGn} on \texttt{SacS} dataset. For traffic flow prediction, \textsc{LocaleGn} outperforms \textsc{RGn} and \textsc{AGn} on all datasets regarding the MAPE, but its performance is slightly lower than \textsc{RGn} on \texttt{SacF} dataset in terms of MAE and RMSE. On \texttt{SF} , the differences between \textsc{AGn} and \textsc{LocaleGn} for MAE and RMSE are within the range of  $\pm$  0.01 and negligible, but MAPE of \textsc{LocaleGn} is substantially smaller than that of \textsc{AGn}. Overall, \textsc{LocaleGn} has the best and most stable performance among the three proposed models on both traffic speed and flow prediction tasks.  The detailed implementation for all the three models is also available in the code repository.

\bibliographystyle{IEEEtran}
\bibliography{ref}
\begin{IEEEbiography}
[{\includegraphics[width=1in,height=1.25in,clip,keepaspectratio]{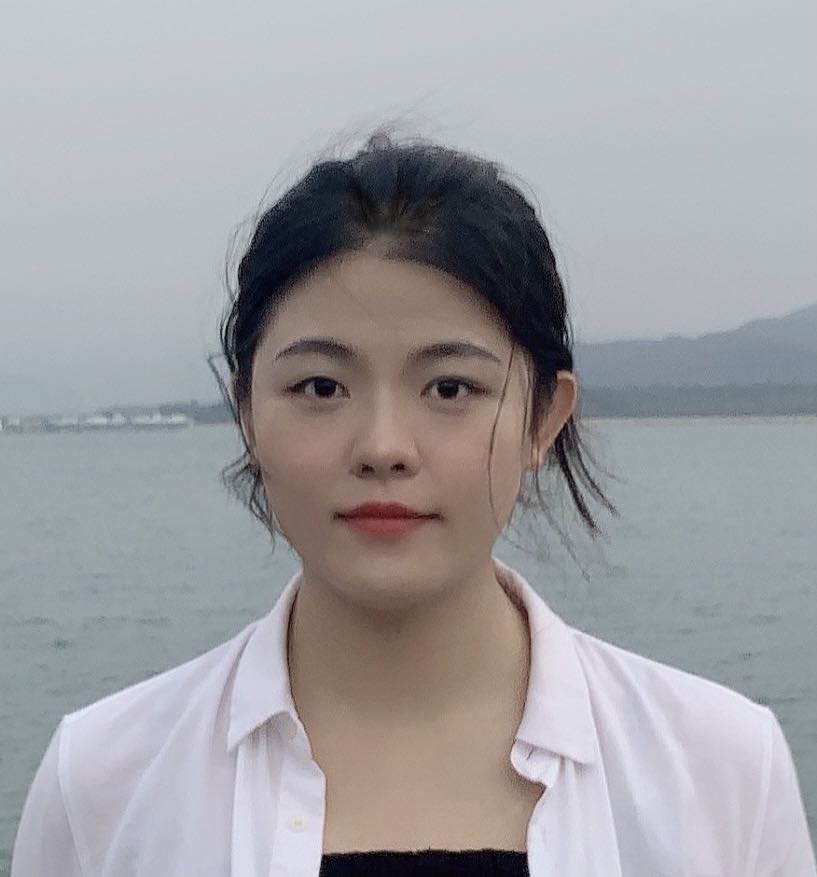}}]{Mingxi Li} graduated from Sichuan University, and she is currently a Ph.D. student with the Department of Civil and Environmental Engineering at the Hong Kong Polytechnic University (PolyU). Her research interests include deep learning, multi-source traffic data mining, urban computing, and the corresponding applications in intelligent transportation systems (ITS).
\end{IEEEbiography}

\begin{IEEEbiography}[{\includegraphics[width=1in,height=1.25in,clip,keepaspectratio]{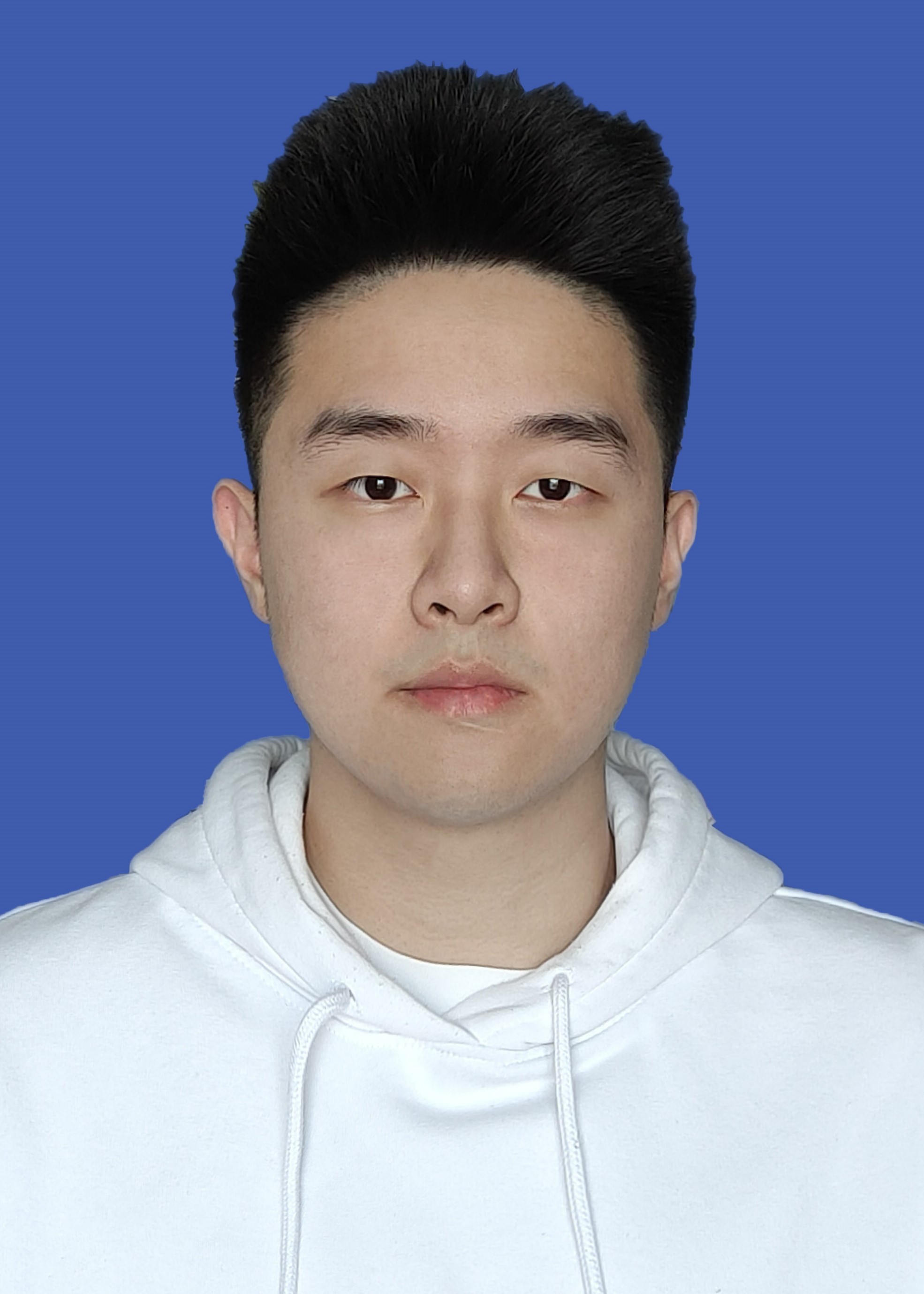}}]{Yihong Tang} received a bachelor's degree in Computer Science and Technology from the Beijing University of Posts and Telecommunications (BUPT), and he is now a Master of Philosophy (MPhil) student at the Department of Urban Planning and Design, University of Hong Kong (HKU). His research interests include data and graph mining, urban computing, demand and mobility modeling, privacy and security issues in emerging internet of things systems, and intelligent transportation applications.
\end{IEEEbiography}

\begin{IEEEbiography}[{\includegraphics[width=1in,height=1.25in,clip,keepaspectratio]{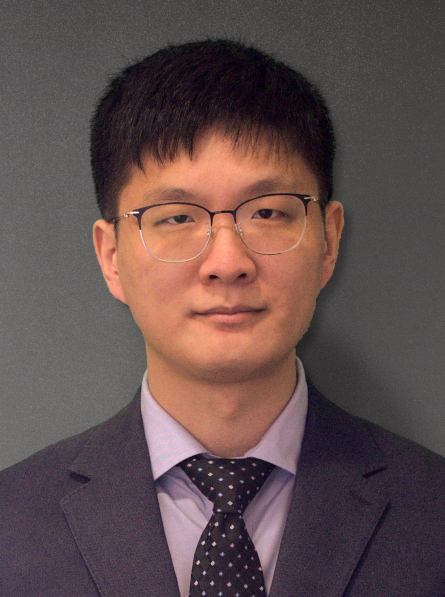}}]{Wei Ma} (IEEE member) received bachelor’s degrees in Civil Engineering and Mathematics from Tsinghua University, China, master degrees in Machine Learning and Civil and Environmental Engineering, and PhD degree in Civil and Environmental Engineering from Carnegie Mellon University, USA. He is currently an assistant professor with the Department of Civil and Environmental Engineering at the Hong Kong Polytechnic University (PolyU). His research focuses on intersection of machine learning, data mining, and transportation network modeling, with applications for smart and sustainable mobility systems. He has received 2020 Mao Yisheng Outstanding Dissertation Award and best paper award (theoretical track) at INFORMS Data Mining and Decision Analytics Workshop. Dr. Ma is now serving in the Early Career Editorial Advisory Board on Transportation Research Part C: Emerging Technologies.
\end{IEEEbiography}

\end{document}